\pdfoutput=1
\documentclass[10pt,twocolumn,letterpaper]{article}

\usepackage[pagenumbers]{cvpr}

\usepackage{cuted} 

\usepackage{currfile} 

\usepackage{caption} 
\usepackage{flafter} 

\usepackage{booktabs,multirow,amsmath,amssymb,amsthm,xcolor,graphicx,bm}
\newtheorem{proposition}{Proposition}
\newcommand{\method}{TASP}

\definecolor{cvprblue}{rgb}{0.21,0.49,0.74}
\usepackage[pagebackref,breaklinks,colorlinks,allcolors=cvprblue]{hyperref}
\def\confName{CVPR}
\def\confYear{2027}
\def\arxivversion{1}

\newcommand{\NumBtnBestAll}{84.4}
\newcommand{\NumBtnBestClean}{82.7}

\newcommand{\NumBtnClipDinoHm}{80.9}
\newcommand{\NumBtnClipDinoNovel}{74.3}

\newcommand{\NumBtnCoopPeHm}{85.1}
\newcommand{\NumBtnCoopPeNovel}{81.3}

\newcommand{\NumBtnKgcoopPeHm}{87.7}
\newcommand{\NumBtnKgcoopPeNovel}{86.8}

\newcommand{\NumBtnPeDinoHm}{88.3}

\newcommand{\NumBtnPeHm}{87.8}
\newcommand{\NumBtnPeLBase}{92.7}

\newcommand{\NumBtnPeLHm}{90.8}
\newcommand{\NumBtnPeLNovel}{88.9}
\newcommand{\NumBtnPeNovel}{85.8}

\newcommand{\NumBtnPromptsrcPeHm}{88.2}
\newcommand{\NumBtnPromptsrcPeNovel}{86.3}

\newcommand{\NumClapRealGapMaxPeL}{2.4}

\newcommand{\NumClapRealGapMinPeL}{1.7}

\newcommand{\NumClapRealPeLOne}{84.7}
\newcommand{\NumClapRealPeLSixteen}{89.4}

\newcommand{\NumDgBestTgt}{61.6}

\newcommand{\NumDgClipDinoTgt}{63.0}

\newcommand{\NumDgPeDinoTgt}{71.9}

\newcommand{\NumDgPeLDinoTgt}{81.8}

\newcommand{\NumDgPeLTgt}{82.8}

\newcommand{\NumFDinoOne}{62.0}

\newcommand{\NumFDinoEight}{83.1}
\newcommand{\NumFDinoSixteen}{86.2}

\newcommand{\NumFClipDinoOne}{74.8}
\newcommand{\NumFClipDinoSixteen}{87.2}

\newcommand{\NumFClipOne}{71.1}
\newcommand{\NumFClipSixteen}{81.4}

\newcommand{\NumFPeDinoOne}{83.2}
\newcommand{\NumFPeDinoSixteen}{89.2}

\newcommand{\NumFPeLEight}{90.3}
\newcommand{\NumFPeLFour}{89.4}
\newcommand{\NumFPeLOne}{87.1}
\newcommand{\NumFPeLSixteen}{91.4}

\newcommand{\NumFPeLStdMax}{0.2}

\newcommand{\NumFPeLTwo}{88.2}
\newcommand{\NumFPeOne}{82.7}
\newcommand{\NumFPeSixteen}{88.0}

\newcommand{\NumFreqClipHigh}{36.7}
\newcommand{\NumFreqClipHighRel}{49}

\newcommand{\NumFreqDinoHigh}{56.1}
\newcommand{\NumFreqDinoHighRel}{65}

\newcommand{\NumFreqDinoTwoHigh}{49.5}
\newcommand{\NumFreqDinoTwoHighRel}{60}

\newcommand{\NumFreqSigHigh}{44.7}
\newcommand{\NumFreqSigHighRel}{54}

\newcommand{\NumGainMaxPeDino}{7.1}
\newcommand{\NumGainMaxPeL}{11.0}

\newcommand{\NumGainMinPeDino}{2.4}
\newcommand{\NumGainMinPeL}{4.6}

\newcommand{\NumKappaAccHatClip}{70.9}

\newcommand{\NumKappaAccImgClip}{67.9}

\newcommand{\NumKappaAccValClip}{73.8}

\newcommand{\NumKappaValClip}{4}
\newcommand{\NumKappaValClipL}{4}
\newcommand{\NumKappaValMeta}{8}
\newcommand{\NumKappaValPe}{8}
\newcommand{\NumKappaValPeL}{8}
\newcommand{\NumKappaValSL}{1}
\newcommand{\NumKappaValSig}{4}
\newcommand{\NumKappaValSigTwo}{4}

\newcommand{\NumKshotMaxDiff}{0.3}

\newcommand{\NumLodoLossPeDino}{0.5}

\newcommand{\NumLoraMinClipSixteen}{30}
\newcommand{\NumLoraMinPeLSixteen}{52}
\newcommand{\NumLoraMinPeSixteen}{49}

\newcommand{\NumLoraParamsMax}{0.59}
\newcommand{\NumLoraParamsMin}{0.18}

\newcommand{\NumLoraPeakPeL}{24.3}

\newcommand{\NumPubBestEight}{83.8}
\newcommand{\NumPubBestFour}{80.6}
\newcommand{\NumPubBestOne}{76.1}
\newcommand{\NumPubBestSixteen}{86.8}
\newcommand{\NumPubBestTwo}{77.9}
\newcommand{\NumPubCurves}{86}

\newcommand{\NumPubPapers}{74}

\newcommand{\NumSelLamClipDino}{10^{-5}}

\newcommand{\NumSelLamPeDino}{10^{-3}}
\newcommand{\NumSelLamPeL}{10^{-3}}

\newcommand{\NumSelWClipDino}{16}

\newcommand{\NumSelWPeDino}{1}
\newcommand{\NumSelWPeL}{1}

\newcommand{\NumSrcCollapsed}{20}
\newcommand{\NumSrcExcluded}{14}

\newcommand{\NumSslGainClipSixteen}{6.5}

\newcommand{\NumSslGainPeLSixteen}{0.6}

\newcommand{\NumSslGainPeSixteen}{1.9}

\newcommand{\NumSslGainSigSixteen}{3.3}

\newcommand{\NumSslGainSigTwoSixteen}{2.4}

\newcommand{\NumTascGapPeLOne}{1.8}
\newcommand{\NumTascGapPeLSixteen}{1.3}

\newcommand{\NumTimeImnetPeL}{4.5}

\newcommand{\NumTimeMeanPeL}{0.1}

\newcommand{\NumTrGainPeDinoOne}{1.4}

\newcommand{\NumTrGainPeLOne}{1.0}

\newcommand{\NumTrSelPeDino}{$\beta=3$, $t_0=0.3$, 1 iteration}
\newcommand{\NumTrSelPeL}{$\beta=3$, $t_0=0.3$, 1 iteration}
\newcommand{\NumValOwnPeDino}{87.3}

\newcommand{\NumValVisPeDino}{87.1}

\newcommand{\NumXValPeL}{90.3}
\newcommand{\NumXValPeLDino}{90.5}

\newcommand{\NumZsClipPub}{65.1}

\newcommand{\NumZsGainPeLOne}{3.3}
\newcommand{\NumZsGainPeLSixteen}{7.6}

\newcommand{\NumZsStdClipEleven}{65.3}

\newcommand{\NumZsStdPeEleven}{79.7}
\newcommand{\NumZsStdPeLEleven}{83.8}

\title{Text Is Just Another Shot in Vision-Language Models:\\
Simple Probes on Frozen Encoders Do It}

\author{
Reda Bensaid$^{1,3}$ \quad
Ismail Ben Ayed$^{2}$ \quad
Fran\c{c}ois Leduc-Primeau$^{1}$ \quad
Vincent Gripon$^{3}$\\[0.4em]
$^{1}$Polytechnique Montr\'eal, Montr\'eal, Canada\\
$^{2}$LIVIA, ILLS, ETS Montr\'eal, Canada\\
$^{3}$IMT Atlantique, Brest, France
}

\begin{document}
\maketitle
\begin{abstract}
Few-shot adaptation of vision-language models has produced increasingly elaborate methods built on CLIP and on one
shared principle: the zero-shot text classifier is the anchor, and labeled images only correct it. We argue that this
principle reflects CLIP's visual features rather than a property of the task. We provide a simple Bayesian interpretation, in which a class name is
worth exactly $\kappa=\sigma^2/\tau^2$ labeled images, the ratio between the visual spread of a class and the
misalignment of its name. The Bayes classifier of this model counts text as $\kappa$ more samples. With modern frozen
encoders, few-shot adaptation then reduces to a well-regularised convex probe. Specifically, we introduce TASP, a simple convex logistic regression on frozen features that incorporates class-name embeddings as weighted training examples and as a regularization prior. 
On the standard 11-dataset benchmark, TASP with frozen PE-Core-L features achieves \NumFPeLOne\% accuracy at 1 shot and \NumFPeLSixteen\% at 16 shots, outperforming state-of-the-art few-shot vision-language adaptation methods by 
$+$\NumGainMinPeL{} to $+$\NumGainMaxPeL{} percentage points. While stronger representations explain much of this improvement, controlled comparisons on matched features show that existing frozen-encoder adaptation methods do not materially improve on this probe. We further find that the weight assigned to text matters more than how it is incorporated, and that combining self-supervised visual features with vision-language representations is most beneficial for weaker visual encoders. Together, these findings suggest that progress in few-shot vision-language adaptation depends more on representation quality and the appropriate use of text supervision than on increasingly complex adaptation mechanisms.

\end{abstract}

\section{Introduction}
\label{sec:intro}

\begin{figure*}[t]
\centering
\includegraphics[width=\textwidth]{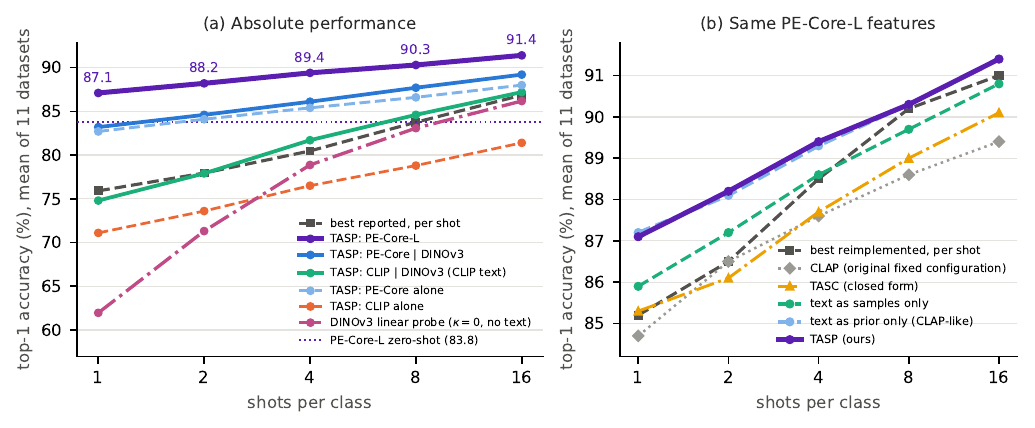}
\caption{\textbf{Strong features explain the absolute gains; the probe remains competitive under controlled features.}
(a) Mean accuracy over the 11 datasets of the few-shot benchmark. Black dashed: the best reported value
at each shot count among the \NumPubCurves{} method and baseline curves reporting all 11 datasets with CLIP ViT-B/16,
including arXiv preprints and methods using LLM-generated text. Coloured: \method{}, a single logistic regression on
frozen features in which each class name is one more training sample, with one hyper-parameter configuration per feature
set for all datasets and shot counts, selected without ImageNet; pink dash-dot: DINOv3 linear probe without text
($\kappa=0$); dotted purple: zero-shot classifier of PE-Core-L. (b) Controlled comparison on identical frozen PE-Core-L
features, splits and model-selection protocol (11 datasets; ImageNet excluded from model selection): \method{}, its samples-only, prior-only and closed-form
variants, CLAP with its originally reported configuration, and the best reimplemented prior method at each shot count. We use the same splits, seeds and shots as the literature.
ImageNet is included in evaluation but never used for hyperparameter selection (Sec.~\ref{sec:setup}). Full controlled results: \cref{tab:samefeat}.}
\label{fig:teaser}
\end{figure*}

Few-shot adaptation of vision-language models has become a crowded field. Starting from CLIP's zero-shot
classifier~\cite{radford2021clip}, the existing methods learn prompts~\cite{zhou2022coop,zhou2022cocoop,
khattak2023maple,khattak2023promptsrc}, add feature adapters and caches of the support
set~\cite{gao2024clipadapter,zhang2022tip,zhu2023ape}, regularize linear probes towards the text
classifier~\cite{silva2024clap,huang2024lp++}, cast the problem as kernel regression around the zero-shot
predictor~\cite{bendou2025proker}, or fine-tune the encoders with low-rank updates~\cite{zanella2024cliplora,
farina2025twostage}. Despite their diversity, these methods share one design principle: \emph{the text is the anchor and
the images are a correction}.

This principle is well founded \emph{for CLIP}. The CLIP authors observed that the zero-shot classifier matches a
4-shot linear probe on the same features and that moving from zero-shot to one or two labeled images can
\emph{decrease} accuracy~\cite{radford2021clip}. With such an encoder, a class name is worth several images. But this is a statement about one encoder, whose image features miss fine-grained and spatial
detail~\cite{tong2024eyes,elbanani2024probe3d}. Self-supervised encoders such as DINOv2/v3~\cite{oquab2024dinov2,
simeoni2025dinov3} cluster classes far more sharply, but have no text head, or only a weak one.

\textbf{We revisit few-shot vision-language classification from the two quantities that matter: how sharply the visual
features cluster each class, and how well the class name is aligned with its images.} We propose a simple Bayesian model
interpretation (Sec.~\ref{sec:method}), which  shows that the posterior treats the class name as exactly $\kappa=\sigma^2/\tau^2$ additional
images, where $\sigma^2$ measures the within-class visual spread and $\tau^2$ the misalignment of the class name with
its images. Under this model, the role of text is therefore governed by its \emph{weight} rather than by a particular
adaptation architecture, and that weight depends on the encoders. This leads us to a deliberately minimal method, \method{} (Text-As-Shot Probe): a convex
logistic regression in which each class name is both a weighted training sample and the centre of the prior on its
class weights, solved to numerical optimality, with one configuration for all datasets and shot counts. When the vision-language model
(VLM) clusters classes poorly, the same probe can take the \emph{union} of its features with those of a frozen
self-supervised (SSL) encoder, with the class names living in the aligned block and the loss deciding which block to trust.

On frozen PE-Core-L~\cite{bolya2025pe} features, \method{} is above every previously reported result on the standard 11-dataset
benchmark at every shot count, by \NumGainMinPeL{} to \NumGainMaxPeL{} points over the best of them (\cref{fig:teaser})
with no fine-tuning. With ViT-B encoders, the size of the CLIP backbone in the literature, the margin is still
\NumGainMinPeDino{} to \NumGainMaxPeDino{} points. A stronger encoder explains most of this, and that is our main point: 
the zero-shot classifier of PE-Core-L alone is already at the level of the best published methods, and the adaptation
mechanisms of the literature, designed around CLIP's weak visual features, barely improve on a convex probe once the
features are good. The controlled comparison in \cref{fig:teaser}b makes the separation explicit: on identical frozen
PE-Core-L features, \method{} and its CLAP-like special case nearly overlap, while the best reimplemented prior
method at each shot count remains below them. Even without a text head, a DINOv3 linear probe reaches \NumFDinoEight\% at 8 shots and
\NumFDinoSixteen\% at 16 shots, already matching the reported few-shot frontier. This is not surprising asymptotically;
what is striking is that it happens within the few-shot regime. When compared to the zero-shot classifier of the same encoder, \method{} reaches $+$\NumZsGainPeLOne{}
to $+$\NumZsGainPeLSixteen{} points. Sec.~\ref{sec:fair} tests this
on matched features and backbones; methods that fine-tune the encoders, a heavier category, are compared in the appendix.

\paragraph{Contributions.}
\begin{itemize}
\item A Bayesian account of multimodal few-shot learning in which the value of a class name is measured in \emph{shots},
$\kappa=\sigma^2/\tau^2$, used as a motivation and measured with the posterior classifier of the model: 4--8 shots for
contrastive VLMs, about one for a lightweight text head (Sec.~\ref{sec:analysis}).
\item The finding that, with modern frozen encoders, a well-regularised convex probe is all few-shot adaptation needs:
a new state of the art at all shot counts, including with ViT-B-sized encoders comparable to those used in prior work,
using a single configuration and only seconds of training (Sec.~\ref{sec:results}).
\item An extensive fairness study (Sec.~\ref{sec:fair}): on identical PE-Core-L and PE-Core$\|$DINOv3 features,
\method{} remains at or above the reimplemented prior methods, while a CLAP-like probe selected like ours nearly
matches it; a self-supervised partner helps in proportion to what the VLM lacks.
\item Evidence that how the text enters the classifier matters little: text as samples and text as prior are
interchangeable along a ridge of equal accuracy, as Proposition~\ref{prop:kappa} predicts; at the $\kappa=0$ endpoint,
DINOv3 without text reaches the reported frontier with only 8 shots.
\end{itemize}

\section{Related work}
\label{sec:related}

\paragraph{Few-shot adaptation of CLIP.}
Despite their different parameterisations, most methods retain the zero-shot classifier as an initialisation,
regulariser or additive anchor. Prompt learning optimises context tokens on the support set~\cite{zhou2022coop,
zhou2022cocoop}, often regularised towards the hand-crafted prompt~\cite{yao2023kgcoop,zhu2023prograd,
khattak2023promptsrc}, with several prompts~\cite{chen2023plot,lafon2024gallop} or in both encoders~\cite{khattak2023maple,
guo2025mmrl}. Adapters and caches add residuals to the zero-shot logits~\cite{gao2024clipadapter,zhang2022tip,zhu2023ape,
yu2023taskres}; training-free methods add Gaussian discriminant logits~\cite{wang2024gda} or fit kernel regression around
the zero-shot predictor~\cite{bendou2025proker}. Linear-probe variants initialise or regularise the probe with text
weights~\cite{silva2024clap}, blend text and visual prototypes~\cite{huang2024lp++}, or add class-name embeddings to its
training set~\cite{lin2023crossmodal}. The latter is the text-as-samples special case of \method{}. Heavier alternatives
fine-tune the visual projector~\cite{fahes2026prolip}, use low-rank encoder updates~\cite{zanella2024cliplora,
farina2025twostage}, or add LLM descriptions~\cite{zhu2024awt}. Our question is not how to build another correction around CLIP, but how much weight its text should
retain when the visual representation changes.

\paragraph{Visual quality and additional encoders.}
CLIP's image features have documented blind spots~\cite{tong2024eyes,elbanani2024probe3d}; self-supervised features help
vision-centric tasks~\cite{tong2024cambrian,fan2025webssl}, and recent vision-language encoders narrow the gap with better
objectives, data and dense features~\cite{tschannen2025siglip2,bolya2025pe}. Prior combinations preserve the same CLIP
anchor: CaFo adds a DINO cache, GPT-3 prompts and DALL-E images to a Tip-Adapter residual~\cite{zhang2023cafo}; AMU-Tuning
adds an uncertainty-weighted bias from MoCo-v3 features~\cite{tang2024amu}; and a Gaussian-process ensemble treats DINO
features as a residual around the zero-shot mean~\cite{miao2024bayesian}. Thus SSL features remain a correction rather
than a first-class representation. Prior inductive methods on this benchmark predominantly use older SSL encoders such
as DINO or MoCo-v3; we evaluate modern DINOv2/v3 and concatenate their features symmetrically with the VLM. We re-implement
CaFo and AMU-Tuning on exactly the same features (\cref{tab:samefeat}). Language-free training on MetaCLIP's
data~\cite{fan2025webssl} provides a data-matched control; because train--test similarity in pretraining data can inflate
generalisation~\cite{mayilvahanan2024traintest,udandarao2024nozeroshot}, we also separate results by the provenance of the
visual encoder.

\paragraph{Aligning text to frozen visual encoders.}
LiT~\cite{zhai2022lit}, ASIF~\cite{norelli2023asif}, ShareLock~\cite{ruthardt2026sharelock}, dino.txt~\cite{jose2025dinotxt}
and SAIL~\cite{zhang2025sail} align language to frozen image encoders. This line of work shows that visual quality and
language alignment need not be learned jointly: a strong self-supervised representation can acquire zero-shot
classification capabilities without changing its visual geometry. It therefore provides a controlled setting for our
model, which separates within-class visual spread $\sigma^2$ from text--image misalignment $\tau^2$. We use ShareLock as
a lightweight alignment head on DINOv2-L to study how weaker language alignment changes the number of shots a class name
is worth.

\paragraph{Simple baselines and Bayesian shrinkage.}
In classic few-shot learning feature quality dominates the adaptation algorithm~\cite{chen2019closer,wang2019simpleshot,
tian2020rethinking,dhillon2020baseline,hu2022pmf,luo2023closer}. Our analysis connects this observation to a standard
Bayesian interpretation: auxiliary observations can be assigned a fractional sample weight~\cite{gelman2013bda,
ibrahim2000power}, while pretrained parameters can define the centre of a regularisation prior~\cite{li2018l2sp}. In our
setting the class name plays both roles: it is a weighted pseudo-example and the centre of the zero-shot classifier.
\method{} combines these two familiar forms of shrinkage and expresses their strength in the common unit of labeled
images.

\section{Text as a shot}
\label{sec:method}

\paragraph{Setting.}
We consider $C$ classes, each represented by a class name and $n$ labeled images (the \emph{support}). A frozen image
encoder $f$ maps images to $\mathbb{R}^D$ and a text encoder $g$ maps the class prompt to the same space, $t_c = g(\text{``a photo of a \{class\}.''})$. Image features are centered on the mean of the support and texts on their
own mean, which removes the modality offset~\cite{liang2022mind}, and both are $\ell_2$-normalized.

\subsection{How many shots is a class name worth?}
\label{sec:kappa}
Consider the simplest generative model of the two sources of evidence about class $c$. Images scatter around an unknown
class mean $m_c$, and the text is a noisy, misaligned observation of the same mean:
\begin{equation}
x \mid c \sim \mathcal{N}(m_c, \sigma^2 I), \qquad t_c \sim \mathcal{N}(m_c, \tau^2 I).
\label{eq:model}
\end{equation}
Here $\sigma^2$ measures how tightly the visual encoder clusters a class, and $\tau^2$ how far the
alignment places the class name from where its images lie.

\begin{proposition}[Text-shot equivalence]
\label{prop:kappa}
Under~\eqref{eq:model} with a flat prior on $m_c$, the posterior mean of $m_c$ given $n$ shots with average
$\bar{x}_c$ and the text $t_c$ is
\begin{equation}
\hat m_c = \frac{n\,\bar{x}_c + \kappa\, t_c}{n + \kappa}, \qquad \kappa = \frac{\sigma^2}{\tau^2}.
\label{eq:posterior}
\end{equation}
The class name is therefore \emph{exactly} equivalent to $\kappa$ additional labeled images, and its share of the
decision, $\kappa/(n+\kappa)$, decays as $1/n$. Equivalently, $t_c$ is the mean of a Gaussian prior on $m_c$ whose
precision equals that of $\kappa$ images.
\end{proposition}
The proof is the standard conjugate update
(\ifdefined\arxivversion\cref{app:proof}\else Appendix A\fi).
Three consequences stem from this and frame the paper.
\textbf{(i) Zero-shot and linear probing are the two ends of one estimator}: $n=0$ gives the zero-shot classifier, and
$n\gg\kappa$ corresponds to a purely visual one. \textbf{(ii) For a given task distribution, $\kappa$ is primarily a property of the encoders.} A visual encoder
that clusters classes poorly makes the text worth many shots, and a sharp one makes it worth a few unless the alignment is
excellent. For CLIP, the zero-shot classifier matches a 4-shot probe~\cite{radford2021clip}, \ie $\kappa\approx4$, which
explains why the literature anchors on the text. \textbf{(iii) The class name can enter as samples or as a prior.} The two
readings of~\eqref{eq:posterior} give the same estimator, so what matters is the weight of the text, not the mechanism.

\subsection{TASP: a text-as-shot linear probe}
\label{sec:tasp}
The discriminative counterpart of~\eqref{eq:posterior} is a multinomial logistic regression that incorporates class
names both as weighted training examples and as a regularization prior. Let $x_{cjr}$ denote view $r\in\{1,\ldots,R\}$
of support image $j\in\{1,\ldots,n\}$ from class $c$, and let $w$ be the effective number of samples assigned to each
class name. With $W\in\mathbb{R}^{D\times C}$, $b\in\mathbb{R}^C$, cross-entropy $\ell$, regularization strength
$\lambda_0$, and zero-shot classifier $W_0=s[t_1,\dots,t_C]$ ($s$ is the usual logit scale), define the loss of one
support image as the average over its views,
$\ell_{cj}^{\rm img}=R^{-1}\sum_{r=1}^R\ell(W^\top x_{cjr}+b,c)$, and the class-name loss as
$\ell_c^{\rm txt}=\ell(W^\top t_c+b,c)$. We optimize
\begin{equation}
\mathcal{L}(W,b)=\frac{\displaystyle\sum_{c=1}^{C}
 \left[\sum_{j=1}^{n}\ell_{cj}^{\rm img}+w\,\ell_c^{\rm txt}\right]}{C(n+w)}
 +\frac{\lambda_0}{2n}\|W-W_0\|_F^2 .
\label{eq:tasp}
\end{equation}
The average over the $R$ augmented views prevents augmentation multiplicity from changing the weight of an image.
Consequently, the two terms inside the brackets give image and text evidence total weights $n$ and $w$ per class:
each class name acts as $w$ additional labeled images (a power prior~\cite{ibrahim2000power}). The last term regularizes
the classifier toward the zero-shot weights, with a strength that decreases as the number of shots increases.

The objective is convex (strictly in $W$ for $\lambda_0>0$),
and is solved to numerical optimality using full-batch L-BFGS, without learning-rate schedules or early stopping. Only two hyper-parameters remain, $w$ and $\lambda_0$, both shared across datasets and shot counts. Special cases include a standard linear probe ($w=0$, $\lambda_0=0$), cross-modal adaptation~\cite{lin2023crossmodal} ($w=n$, $\lambda_0=0$), which equally weights image and text losses, and a CLAP-like probe~\cite{silva2024clap} ($w=0$, while $\lambda_0$ corresponds to Lagrange multipliers). Consequence~(iii) predicts a trade-off between $w$ and $\lambda_0$, which we observe in Sec.~\ref{sec:analysis}.

\subsection{A closed-form counterpart}
\label{sec:tasc}
Replacing the cross-entropy of~\eqref{eq:tasp} by a squared loss on one-hot targets $e_y$, with the same text weight $w$
and the same prior centred on the zero-shot classifier $W_0=[t_1,\dots,t_C]$ (unit logit on its own name), gives a
closed form. To make every average explicit, define, for any function $h$ of a feature--target pair,
\begin{equation*}
\begin{aligned}
\mathbb E_{\rm img}[h]
&=\frac{1}{CnR}\sum_{c=1}^{C}\sum_{j=1}^{n}\sum_{r=1}^{R}h(x_{cjr},e_c),\\
\mathbb E_{\rm txt}[h]
&=\frac{1}{C}\sum_{c=1}^{C}h(t_c,e_c).
\end{aligned}
\end{equation*}
The least-squares objective is
\begin{equation*}
\begin{aligned}
\min_W\quad
&\frac{n}{n+w}\,\mathbb E_{\rm img}[\|W^\top x-e\|^2]\\
&+\frac{w}{n+w}\,\mathbb E_{\rm txt}[\|W^\top t-e\|^2]
+\frac{\lambda_0}{n}\|W-W_0\|_F^2,
\end{aligned}
\end{equation*}
and its solution is
\begin{equation*}
W=(A+\mu I)^{-1}(B+\mu W_0),
\end{equation*}
where
\begin{equation*}
\begin{aligned}
A&=\frac{n\,\mathbb E_{\rm img}[xx^\top]+w\,\mathbb E_{\rm txt}[tt^\top]}{n+w},\\
B&=\frac{n\,\mathbb E_{\rm img}[xe^\top]+w\,\mathbb E_{\rm txt}[te^\top]}{n+w},
\qquad \mu=\lambda_0/n.
\end{aligned}
\end{equation*}
Thus augmented views are averaged within each of the $n$ images, exactly as in~\eqref{eq:tasp}. Features are centred on
the support mean, so a class-independent bias does not change the decision. We call this variant TASC and select
$(w,\lambda_0)$ as for \method{}.

\subsection{Union of encoders}
\label{sec:union}
Proposition~\ref{prop:kappa} separates two qualities that a single model rarely offers at once: a sharp visual space
(small $\sigma^2$, what matters once images are available) and a well-aligned text head (small $\tau^2$, what matters at
$n\approx0$). Contrastive VLMs have the second and, to a degree that grows with their scale and data, the first;
self-supervised encoders (DINOv2/v3) have the first only. \method{} combines them without new machinery. Each image is represented by the concatenation of its features from both
encoders, $x=[f_{\text{VLM}}(\text{img});\,f_{\text{SSL}}(\text{img})]$, each block centred and normalised. Each class name
is represented in the aligned block only, $t_c=[g_{\text{VLM}}(c);\,0]$, and $W_0=s\,[t_1,\dots,t_C]$ accordingly. The same
problem~\eqref{eq:tasp} is solved on the union. The decision $W^\top x$ is a sum of a linear classifier per block. With few images only the aligned block can fit
the class names and the shots at once, so it carries the decision; as $n$ grows the visual likelihood dominates and the
classifier draws on the block that separates classes best.

\section{Experimental setup}
\label{sec:setup}

\paragraph{Benchmark.}
We use the standard few-shot benchmark of CoOp~\cite{zhou2022coop}: ImageNet, Caltech101, OxfordPets, StanfordCars,
Flowers102, Food101, FGVCAircraft, SUN397, DTD, EuroSAT and UCF101, with the official splits, $n\in\{1,2,4,8,16\}$ shots
per class and 3 seeds. The support images are drawn with the seeded sampler of CoOp, which we verified reproduces the
literature's shots exactly. For ImageNet the sampler draws from the file list, whose order CoOp does not fix; we use the
sorted list. As in prior work~\cite{zhang2022tip,zhou2022coop}, each support image is represented by 10 random augmentations
(random resized crop and flip). We also report base-to-novel generalisation~\cite{zhou2022cocoop} in
\ifdefined\arxivversion\cref{app:b2n}\else Appendix C\fi
and robustness to
distribution shift on ImageNet-V2, -Sketch, -A and -R~\cite{recht2019imagenetv2,wang2019imagenetsketch,
hendrycks2021imageneta,hendrycks2021imagenetr}.

\paragraph{Encoders.}
All encoders are frozen and every image is encoded once at $224{\times}224$ (PE-Core-L at its native 336). Vision-language models: CLIP ViT-B/16 and
ViT-L/14 (OpenAI, with QuickGELU as in the original models\footnote{Loading the OpenAI weights into open\_clip's default
\texttt{ViT-B-16} configuration silently replaces QuickGELU with GELU and, in our measurements, costs up to 16 points of zero-shot accuracy; we use
the \texttt{-quickgelu} configurations, which reproduce the reported zero-shot accuracies.}), MetaCLIP
ViT-B/16~\cite{xu2024metaclip}, SigLIP and SigLIP2 ViT-B/16~\cite{zhai2023siglip,tschannen2025siglip2}, SigLIP2
SO400M/14, and PE-Core ViT-B/16 and ViT-L/14 (PE-Core-L)~\cite{bolya2025pe}, with the CoOp prompt templates.
Self-supervised encoders (CLS token): DINOv3 ViT-B/16~\cite{simeoni2025dinov3}, DINOv2 ViT-B/14 and
ViT-L/14~\cite{oquab2024dinov2}, Web-SSL DINO ViT-L/14~\cite{fan2025webssl}, trained on MetaCLIP's images, and, as
partners without retrieval-curated data, DINO ViT-B/16~\cite{caron2021dino} and MAE ViT-B/16~\cite{he2022mae} trained on
ImageNet-1k and a supervised ImageNet-21k ViT-B/16~\cite{steiner2022augreg}. The released
ShareLock~\cite{ruthardt2026sharelock} maps Llama-3-8B text embeddings onto DINOv2-L/14.

\paragraph{Hyper-parameter selection.}
Every method we fit on our features, ours and re-implemented baselines alike, is selected with the same rule. Each configuration is
evaluated on the validation split of each dataset, validation accuracies are averaged with equal weight over the 10
datasets other than ImageNet, and the best configuration is used \emph{for all datasets and all shot counts}. Because the
ImageNet validation split is its test split, \emph{ImageNet is never used to choose anything}; the same holds for the
shifted test sets. The selected configurations are $(w,\lambda_0)=(\NumSelWPeL,\NumSelLamPeL)$ for
PE-Core-L, $(\NumSelWPeDino,\NumSelLamPeDino)$ for PE-Core$\|$DINOv3 and $(\NumSelWClipDino,\NumSelLamClipDino)$ for
CLIP$\|$DINOv3; they lie on a ridge of near-equal validation accuracy (Sec.~\ref{sec:analysis}). Large validation sets have been
criticised as a hidden source of supervision~\cite{silva2024clap}; with two global hyper-parameters this channel is
narrow: selecting on $n$ validation images per class~\cite{silva2024clap} changes no average by more than
\NumKshotMaxDiff{} points. Leave-one-dataset-out selection, which never sees the evaluated dataset, has no effect for
PE-Core-L and costs \NumLodoLossPeDino{} points for PE-Core$\|$DINOv3. Centring class names on
their own mean is also a validation choice: \NumValOwnPeDino\% vs \NumValVisPeDino\% validation accuracy with the image
mean (PE-Core$\|$DINOv3).

\paragraph{Reported results.}
We compare against numbers \emph{as reported}, transcribed from the tables of \NumPubPapers{} papers and preprints, keeping the protocol each paper declares. Most methods use CLIP
ViT-B/16 and tune hyper-parameters per dataset or on few-shot validation sets; we use them as reported, which generally favours
them. The full table with sources is released with the code.

\paragraph{Implementation.}
\method{} is solved with full-batch L-BFGS in PyTorch on one consumer GPU (RTX 4070). With pre-extracted features, training takes \NumTimeMeanPeL{}\,s on average per dataset and \NumTimeImnetPeL{}\,s on ImageNet at 16 shots (160k augmented views, 1000 classes) for PE-Core-L.

\section{Main results}
\label{sec:results}

\Cref{fig:teaser} and \cref{tab:main} compare \method{} with previously reported methods on the full 11-dataset benchmark;
per-dataset results are reported in \ifdefined\arxivversion\cref{tab:perds}\else Appendix Table~8\fi.

\paragraph{A convex probe on a frozen encoder is above every reported result, at every shot count.}
On PE-Core-L features and with one configuration, \method{} reaches
\NumFPeLOne/\NumFPeLTwo/\NumFPeLFour/\NumFPeLEight/\NumFPeLSixteen\% at 1/2/4/8/16 shots (standard deviation over seeds
at most \NumFPeLStdMax{} points). This is above the best reported value at each shot count
(\NumPubBestOne/\NumPubBestTwo/\NumPubBestFour/\NumPubBestEight/\NumPubBestSixteen\%, an envelope over \NumPubCurves{}
curves that no single method reaches) by \NumGainMinPeL{} to \NumGainMaxPeL{} points.

\begin{table}[t]\centering\small\setlength{\tabcolsep}{3.0pt}
\resizebox{\columnwidth}{!}{\begin{tabular}{lccccc}\toprule
Method & 1 & 2 & 4 & 8 & 16\\\midrule
\multicolumn{6}{l}{\emph{Prior work, CLIP ViT-B/16}}\\
Zero-shot CLIP & \multicolumn{5}{c}{65.1}\\
Linear probe~\cite{radford2021clip} & 45.8 & 58.0 & 68.0 & 74.5 & 78.8\\
CoOp~\cite{zhou2022coop} & 67.6 & 70.8 & 74.0 & 76.8 & 80.0\\
Tip-Adapter-F~\cite{zhang2022tip} & 70.9 & 73.7 & 75.6 & 78.3 & 80.7\\
PromptSRC~\cite{khattak2023promptsrc} & 72.3 & 75.3 & 78.3 & 80.7 & 82.9\\
CLIP-LoRA~\cite{zanella2024cliplora} & 72.5 & 75.5 & 77.4 & 80.3 & 83.0\\
MMRL~\cite{guo2025mmrl} & 72.7 & 75.9 & 79.2 & 81.5 & 84.3\\
Best peer-reviewed, per shot & 76.1 & 77.9 & 80.4 & 82.3 & 84.9\\
Best incl.\ arXiv, per shot & 76.1 & 77.9 & 80.6 & 83.8 & 86.8\\\midrule
\multicolumn{6}{l}{\emph{Ours: frozen encoders, one convex probe, one configuration}}\\
Zero-shot PE-Core (our features) & \multicolumn{5}{c}{79.7}\\
Zero-shot PE-Core-L (our features) & \multicolumn{5}{c}{83.8}\\
DINOv3 linear probe ($\kappa=0$, no text) & 62.0 & 71.3 & 78.9 & 83.1 & 86.2\\
\method{} CLIP & 71.1 & 73.6 & 76.5 & 78.8 & 81.4\\
\method{} CLIP $\|$ DINOv3 & 74.8 & 77.9 & \textbf{81.7} & \textbf{84.6} & \textbf{87.2}\\
\method{} PE-Core & \textbf{82.7} & \textbf{84.1} & \textbf{85.4} & \textbf{86.6} & \textbf{88.0}\\
\method{} PE-Core $\|$ DINOv3 & \textbf{83.2} & \textbf{84.6} & \textbf{86.1} & \textbf{87.7} & \textbf{89.2}\\
\method{} PE-Core-L $\|$ DINOv3 & \textbf{87.1} & \textbf{88.2} & \textbf{89.5} & \textbf{90.5} & \textbf{91.6}\\
\method{} PE-Core-L & \textbf{87.1} & \textbf{88.2} & \textbf{89.4} & \textbf{90.3} & \textbf{91.4}\\
\bottomrule\end{tabular}}\caption{\textbf{Mean accuracy over the 11 datasets} of the few-shot benchmark (official CoOp splits and shots, 3 seeds). Prior-work results are transcribed from the corresponding source tables. Ours: DINOv3 is a linear probe without text; all \method{} rows use one configuration per feature set for all datasets and shot counts. ImageNet is included in evaluation but never used for hyperparameter selection (Sec.~\ref{sec:setup}). Bold: above every reported CLIP result.}\label{tab:main}\end{table}

\paragraph{The encoder does most of it; the probe adds the images.}
The zero-shot classifier of PE-Core-L alone reaches \NumZsStdPeLEleven\% (PE-Core: \NumZsStdPeEleven\%; CLIP:
\NumZsStdClipEleven\%, vs \NumZsClipPub\% reported), at the level of the best reported few-shot methods up to 8 shots.
On top of it, \method{} adds $+$\NumZsGainPeLOne{} points at 1 shot and $+$\NumZsGainPeLSixteen{} at 16. At the ViT-B scale of
the literature, PE-Core alone (\NumFPeOne\% at 1 shot, \NumFPeSixteen\% at 16) and PE-Core$\|$DINOv3 (\NumFPeDinoOne\% and
\NumFPeDinoSixteen\%) are still above the envelope at every shot count. On CLIP itself \method{} is a strong linear probe
(\NumFClipOne\% and \NumFClipSixteen\%) but below the best reported methods, which were designed around its weak visual
features; adding DINOv3 to the \emph{same} CLIP text lifts it to \NumFClipDinoOne\% and \NumFClipDinoSixteen\%, above every
reported CLIP method from 4 shots on.

\paragraph{Text becomes dispensable surprisingly early.}
DINOv3 has no aligned text head and is therefore the $\kappa=0$ endpoint of our model. Its plain linear probe is weak at
one shot (\NumFDinoOne\%), as expected when no class-name prior is available, but reaches
\NumFDinoEight\% at 8 shots and \NumFDinoSixteen\% at 16. This already exceeds the best peer-reviewed result at both
shot counts and comes within one point of the full reported envelope. That visual evidence eventually dominates text is
unsurprising; that it does so with only eight labeled images per class is the finding.

\begin{figure*}[!t]
\centering
\includegraphics[width=.705\linewidth,trim=0 0 139bp 0,clip]{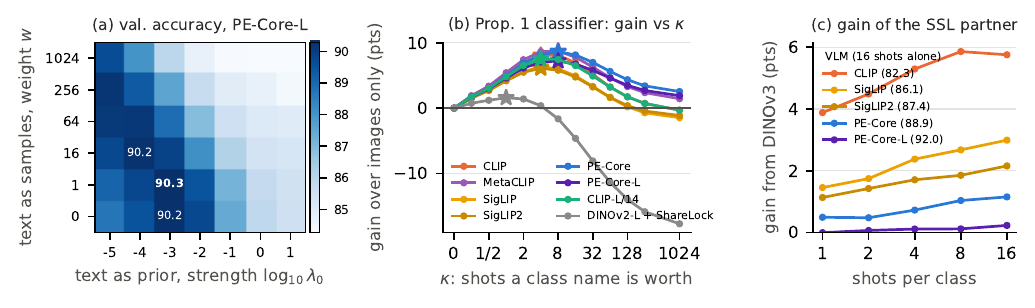}\hfill
\includegraphics[width=.285\linewidth]{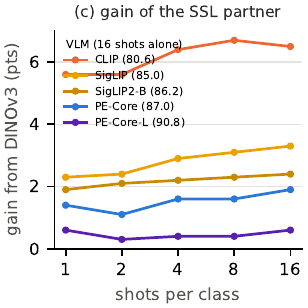}
\caption{\textbf{The weight of the text, not its mechanism.} (a) Validation accuracy of \method{} on PE-Core-L
(10 datasets, all shot counts) over the text weight $w$ (text as samples) and the strength $\lambda_0$ of the
text-centred prior: a ridge of equal accuracy runs from ``prior only'' to ``samples only''. (b) Gain of the Bayes
classifier of Proposition~\ref{prop:kappa} (posterior class means) over images only, as a function of $\kappa$; stars mark
the optimum. A class name is worth about 4--8 images with contrastive vision-language models and about one with the
lightweight ShareLock head. (c) Gain brought by DINOv3 to the text-as-samples probe on each vision-language model (11 datasets; in
parentheses, \method{} on the model alone at 16 shots): the stronger the model alone, the less the partner adds.}
\label{fig:kappa}
\end{figure*}

\paragraph{What a self-supervised partner adds.}
It adds in proportion to what the vision-language model lacks (\cref{fig:kappa}c): at 16 shots,
$+$\NumSslGainClipSixteen{} points for CLIP, $+$\NumSslGainSigSixteen{} for SigLIP, $+$\NumSslGainSigTwoSixteen{} for
SigLIP2, $+$\NumSslGainPeSixteen{} for PE-Core and $+$\NumSslGainPeLSixteen{} for PE-Core-L (11 datasets). With PE-Core-L the
union reaches only \NumXValPeLDino\% vs \NumXValPeL\% validation accuracy, for a second encoder; we therefore
keep PE-Core-L alone as our main configuration, and the union at the ViT-B scale, where it matters. PE-Core is harder to
improve than earlier VLMs for a reason: its authors first develop a heavily regularised, large-scale contrastive image
pretraining recipe, whose internal visual features can rival SSL encoders, before the subsequent language and
spatial alignment stages~\cite{bolya2025pe}. DINOv3 therefore adds partly redundant visual evidence.
The redundancy is strongest with scarce labels: over the 11 datasets, DINOv3 alone trails PE-Core by
20.7 points at 1 shot (\NumFDinoOne\% vs \NumFPeOne\%), but nearly closes the gap by 16
(\NumFDinoSixteen\% vs \NumFPeSixteen\%). In the full TASP configuration of \cref{tab:main}, the gain from their union
grows from $+0.5$ to $+1.2$ points as the support provides enough evidence to exploit the SSL block.

\subsection{Controlled comparisons}
\label{sec:fair}

Our encoders are stronger than CLIP ViT-B/16, so \cref{tab:main} alone does not show that the method matters. We
separate the contributions of the features, the backbone, the second encoder and its pretraining data. The main
controlled accuracy comparisons report all 11 datasets, with ImageNet excluded only from model selection (Sec.~\ref{sec:setup}).

\begin{table}[t]\centering\small\setlength{\tabcolsep}{2.4pt}
\resizebox{\columnwidth}{!}{\begin{tabular}{l|ccc|ccc}\toprule
 & \multicolumn{3}{c}{PE-Core-L} & \multicolumn{3}{c}{PE-Core $\|$ DINOv3}\\
Method (same features) & 1 & 4 & 16 & 1 & 4 & 16\\\midrule
\method{} (ours) & 87.1 & \textbf{89.4} & \textbf{91.4} & \textbf{83.2} & \textbf{86.1} & \textbf{89.2}\\
\quad text as samples only & 85.9 & 88.6 & 90.8 & 82.1 & 85.5 & 88.9\\
\quad text as prior only (CLAP-like~\cite{silva2024clap}) & \textbf{87.2} & 89.3 & \textbf{91.4} & 83.0 & \textbf{86.1} & \textbf{89.2}\\
\quad closed form, least squares (TASC) & 85.3 & 87.7 & 90.1 & 81.4 & 84.7 & 87.5\\
\midrule
CLAP~\cite{silva2024clap} & 84.7 & 87.6 & 89.4 & 80.7 & 84.5 & 86.8\\
Cross-modal LP~\cite{lin2023crossmodal} & 81.0 & 87.3 & 90.7 & 76.6 & 84.0 & 88.6\\
LP++~\cite{huang2024lp++} & 83.0 & 88.0 & 91.0 & 77.2 & 84.6 & 88.8\\
ProKeR~\cite{bendou2025proker} & 85.2 & 87.9 & 90.6 & 80.8 & 84.2 & 88.0\\
GDA~\cite{wang2024gda} & 68.9 & 88.5 & 90.2 & 66.1 & 85.7 & 86.4\\
Tip-Adapter~\cite{zhang2022tip} & 84.5 & 86.9 & 88.6 & 79.9 & 83.3 & 85.7\\
APE~\cite{zhu2023ape} & 84.4 & 86.7 & 88.7 & 79.7 & 83.0 & 86.1\\
Logit ensemble & 84.8 & 87.8 & 90.7 & 80.1 & 85.0 & 88.7\\
\midrule
AMU-Tuning~\cite{tang2024amu} & -- & -- & -- & 79.9 & 82.6 & 86.1\\
CaFo caches~\cite{zhang2023cafo} & -- & -- & -- & 76.1 & 81.7 & 83.8\\
\quad with fine-tuned keys & -- & -- & -- & 79.2 & 84.1 & 87.3\\
\bottomrule\end{tabular}}\caption{\textbf{Prior methods on exactly our features} (mean over all 11 datasets; ImageNet is excluded from model selection). Every method is re-implemented on the same frozen features, with a grid at least as wide as its paper's and the same global selection. AMU-Tuning and CaFo already combine CLIP with a self-supervised encoder; here they receive ours (--: no self-supervised partner in a single-encoder feature set). Bold: best (ties included).}\label{tab:samefeat}\end{table}

\paragraph{F1: Prior methods given exactly our features.}
We re-implement ten methods on the same frozen features and select them with the same rule, over grids at least as wide
as in their papers (CLAP keeps its originally reported, validation-free configuration; \cref{tab:samefeat}). Two of them, AMU-Tuning and CaFo, are the prior methods that already combine
CLIP with a self-supervised encoder; here they receive DINOv3 instead of MoCo-v3 or DINO-v1 (CaFo without its GPT-3
prompts and DALL-E images). In the 11-dataset comparison of \cref{tab:samefeat}, \method{} remains at or above every
prior method on both feature sets, but no single baseline is consistently closest: CLAP, ProKeR, GDA and LP++ alternate
with the shot count.
CLAP~\cite{silva2024clap}, the linear probe regularised towards the zero-shot classifier with class-adaptive
penalties, keeps its fixed originally reported configuration and is \NumClapRealGapMinPeL{} to \NumClapRealGapMaxPeL{} points
below \method{} on PE-Core-L (\NumClapRealPeLOne\% at 1 shot, \NumClapRealPeLSixteen\% at 16). Yet the same
idea with its strength selected like ours, the CLAP-like special case of \method{} (text as prior only), is within a
few tenths of \method{} on both feature sets. Given modern features, a well-regularised convex probe is what matters,
and the regularisation strength more than its form; the text-as-samples term of \method{} is the reading of
Proposition~\ref{prop:kappa}, not a decisive mechanism. Cache and residual designs that anchor
on the zero-shot logits lose the most as shots accumulate (\cref{tab:samefeat}).

\paragraph{Is \method{} training-free?}
Not strictly (\ifdefined\arxivversion\cref{tab:cost}\else Tab.~6\fi)
: it solves a convex problem iteratively, but converges to the same solution across
initialisations and sufficiently large L-BFGS iteration budgets. It takes only seconds and requires neither a manually
chosen learning rate nor an epoch schedule. Training-free methods
trade the logistic loss for a closed form, and lose accuracy for it. TASC, the
least-squares counterpart of \method{} with the same two uses of the class name, is a single linear solve as described in Section~\ref{sec:tasc};
it trails \method{} by \NumTascGapPeLOne{} points at 1 shot and \NumTascGapPeLSixteen{} at 16 on
PE-Core-L (\cref{tab:samefeat}), as Tip-Adapter, APE, GDA and ProKeR, also closed forms, trail it.

\begin{table}[t]\centering\small\setlength{\tabcolsep}{2.5pt}
\resizebox{\columnwidth}{!}{\begin{tabular}{lrrccc}\toprule
Frozen features & M & img/s & 1 & 4 & 16\\\midrule
\multicolumn{6}{l}{\emph{One vision-language model}}\\
CLIP ViT-B/16 & 86 & 1066 & 68.4 & 75.1 & 80.6\\
SigLIP ViT-B/16 & 93 & 1081 & 76.2 & 81.3 & 85.0\\
PE-Core ViT-B/16 & 94 & 942 & 80.7 & 83.9 & 87.0\\
SigLIP2 SO400M/14 & 428 & 196 & 83.5 & 87.6 & 90.1\\
PE-Core ViT-L/14 (336 px) & 317 & 100 & 85.9 & 88.6 & 90.8\\
\multicolumn{6}{l}{\emph{Two vision-language models}}\\
CLIP $\|$ SigLIP & 179 & 537 & 77.4 & 81.6 & 85.5\\
SigLIP2 $\|$ PE-Core & 187 & 504 & 82.1 & 85.1 & 88.1\\
\multicolumn{6}{l}{\emph{Vision-language + self-supervised}}\\
CLIP $\|$ DINOv3 & 172 & 493 & 74.0 & 81.5 & 87.1\\
SigLIP $\|$ Web-SSL$^\ddagger$ & 397 & 216 & 77.4 & 82.4 & 86.9\\
SigLIP $\|$ DINOv3 & 179 & 496 & 78.5 & 84.2 & 88.3\\
PE-Core $\|$ DINOv3 & 180 & 464 & 82.1 & 85.5 & 88.9\\
PE-Core-L $\|$ DINOv3 & 403 & 90 & \textbf{86.5} & \textbf{89.0} & \textbf{91.4} \\
\bottomrule\end{tabular}}\caption{\textbf{What the union needs} (mean over all 11 datasets; the text-as-samples variant of \method{}, rather than the full \method{} used in \cref{tab:main}; one configuration per row, selected without ImageNet). M: vision parameters (millions); img/s: measured encoding throughput on one RTX 4070, fp16 (a union costs the sum of its encoders). $^\ddagger$Web-SSL DINO is trained on MetaCLIP's own images (MC-2B) without retrieval around benchmark images. Other feature sets: \ifdefined\arxivversion\cref{tab:variants}\else Tab.~10\fi. Bold: best in each column.}\label{tab:controls}\end{table}

\paragraph{F2: Second encoders and larger ones.}
\Cref{tab:controls} varies the encoders (text as samples, one configuration per row). At the ViT-B scale, \emph{a
second encoder helps most when it contributes strong visual features}. At 16 shots, DINOv3 raises CLIP from 80.6\% to
87.1\% and SigLIP from 85.0\% to 88.3\%, against a smaller gain from 87.0\% to 88.9\% for PE-Core. A second VLM is less
effective: CLIP$\|$SigLIP reaches 85.5\%, below CLIP$\|$DINOv3, and SigLIP2$\|$PE-Core reaches 88.1\%, below
PE-Core$\|$DINOv3. \emph{A larger modern VLM does better still}: SigLIP2 SO400M reaches 90.1\% and PE-Core-L 90.8\%
without a second encoder; adding DINOv3 to PE-Core-L yields only 91.4\%. The union is thus a way to buy visual quality
at the ViT-B scale, consistent with Proposition~\ref{prop:kappa}: it matters most when the VLM clusters classes poorly.
PE-Core is the difficult case precisely because its contrastive pretraining was engineered to produce unusually strong
general visual features before the PE family's later alignment tuning~\cite{bolya2025pe}; the SSL partner complements a
weaker VLM, but partly duplicates PE-Core's visual competence.

The price of the union is at inference: about half the throughput of one ViT-B (\cref{tab:controls}); PE-Core-L at
336~px costs more than both. Additional matched-backbone, pretraining-data and cost controls are in the appendix.

\subsection{Beyond the standard inductive setting}
\label{sec:gen}

\begin{table}[t]\centering\small\setlength{\tabcolsep}{3pt}
\resizebox{\columnwidth}{!}{\begin{tabular}{lcccccc}\toprule
Method & IN & -V2 & -S & -A & -R & Avg\\\midrule
\multicolumn{7}{l}{\emph{Prior work, CLIP ViT-B/16}}\\
CLIP & 66.7 & 60.8 & 46.1 & 47.8 & 74.0 & 57.2\\
CoOp & 71.5 & 64.2 & 48.0 & 49.7 & 75.2 & 59.3\\
CoCoOp & 71.0 & 64.1 & 48.8 & 50.6 & 76.2 & 59.9\\
MaPLe & 70.7 & 64.1 & 49.1 & 50.9 & 77.0 & 60.3\\
PromptSRC & 71.3 & 64.3 & 49.5 & 50.9 & 77.8 & 60.6\\
Best target avg.\ (TAC) & 72.8 & 66.0 & 50.3 & 51.7 & 78.5 & 61.6\\
\midrule\multicolumn{7}{l}{\emph{Ours: \method{}, 16 shots}}\\
\method{} PE-Core-L $\|$ DINOv3 & \textbf{84.8} & 77.8 & 69.7 & 86.2 & 93.3 & 81.8\\
\method{} PE-Core-L & 84.7 & \textbf{78.2} & \textbf{71.0} & \textbf{88.1} & \textbf{93.9} & \textbf{82.8} \\
\method{} PE-Core $\|$ DINOv3 & 81.4 & 73.3 & 63.9 & 63.8 & 86.6 & 71.9\\
\method{} PE-Core & 79.7 & 71.7 & 62.3 & 62.0 & 85.1 & 70.3\\
\method{} SigLIP $\|$ DINOv3 & 80.1 & 71.5 & 63.7 & 48.5 & 86.8 & 67.6\\
\method{} CLIP $\|$ DINOv3 & 78.0 & 69.1 & 53.4 & 51.2 & 78.3 & 63.0\\
\method{} SigLIP & 78.3 & 69.7 & 65.0 & 46.1 & 87.7 & 67.1\\
\method{} CLIP & 72.1 & 64.1 & 47.4 & 47.6 & 75.0 & 58.5\\
\bottomrule\end{tabular}}\caption{\textbf{Robustness to distribution shift}: trained on 16-shot ImageNet, evaluated unchanged on ImageNet-V2, -Sketch, -A and -R (A, R: their 200 classes); Avg: mean of the four shifted sets. Prior-work rows exclude methods using extra data or test-time adaptation. Bold: best in each column.}\label{tab:dg}\end{table}

\paragraph{Distribution shift.}
\Cref{tab:dg} evaluates the 16-shot ImageNet probe, unchanged, on four shifted test sets. Averaged over the four targets,
\method{} on PE-Core-L reaches \NumDgPeLTgt\% and on PE-Core$\|$DINOv3 \NumDgPeDinoTgt\%, against \NumDgBestTgt\% for the
best reported CLIP method. A SSL partner helps PE-Core on every test set, but lowers PE-Core-L on all four
(PE-Core-L$\|$DINOv3: \NumDgPeLDinoTgt\%), as it lowers SigLIP on sketches and renditions: under large shifts the
vision-language features are the most robust. With CLIP's text, the
union is above every reported CLIP method on average (\NumDgClipDinoTgt\%).

\section{Analysis}
\label{sec:analysis}

\paragraph{A1: How many shots is a class name worth?}
We measure $\kappa$ with the classifier of Proposition~\ref{prop:kappa}, a nearest class mean with posterior
prototypes~\eqref{eq:posterior}, on single encoders with a text head (\cref{fig:kappa}b). The validation-optimal $\kappa$ is \NumKappaValClip{} for CLIP, \NumKappaValSig{} for SigLIP,
\NumKappaValSigTwo{} for SigLIP2, \NumKappaValClipL{} for CLIP-L/14, \NumKappaValMeta{} for MetaCLIP, \NumKappaValPe{} for
PE-Core and \NumKappaValPeL{} for PE-Core-L. For DINOv2-L with the released ShareLock head, whose
text was aligned on 12M captions only, it is \NumKappaValSL: there the text is worth about one image: weighting it
like a CLIP text already costs accuracy, and the text alone falls more than 20 points below images only; for a given task
distribution, the weight of the text is primarily a property of the encoders.

\paragraph{A2: Text as samples or text as prior?}
Proposition~\ref{prop:kappa} reads the class name either as $\kappa$ extra samples or as a prior of the same strength.
\Cref{fig:kappa}a shows the discriminative counterpart: the validation accuracy of \method{} is constant, to within
0.2 point, along a ridge that trades text-as-samples ($w$) for text-as-prior ($\lambda_0$); the same ridge appears on
the unions. Text-anchored and symmetric designs
are two parametrisations of one estimator, which differ by the \emph{weight} they give the text. Even at small $\lambda_0$ the prior
centre fixes the directions the few shots leave unconstrained, so text as samples alone (prior at 0) trails \method{} at 1 shot. This explains \cref{tab:samefeat}: the CLAP-like probe, a point on this ridge, is close to
\method{}; cross-modal adaptation, which ties the text weight to the number of shots ($w=n$), under-weights the text at
1--2 shots; and residual and cache designs, which fix the weight of the zero-shot logits, fall behind as shots accumulate.

\paragraph{A3: What do self-supervised features retain?}
We evaluate the encoders after retaining only image frequencies with magnitude
$|f|\geq 32$ (\ifdefined\arxivversion\cref{app:freq}\else Appendix F\fi).
DINOv3 preserves
\NumFreqDinoHighRel\% of its clean accuracy, compared with
\NumFreqClipHighRel\% for CLIP. This greater retention of high-frequency
visual information provides one possible explanation for the benefits of
DINOv3 on visually detailed datasets such as FGVC-Aircraft and EuroSAT.

\section{Conclusion}
\label{sec:conclusion}

Few-shot vision-language adaptation has largely focused on preserving and refining CLIP's text classifier, whose visual
features are often the limiting factor. Under our Gaussian model, a class name is worth
$\kappa=\sigma^2/\tau^2$ labeled images for a given encoder and task distribution. Empirically, using it as weighted
samples or as the centre of a prior yields nearly identical accuracy when their strengths are selected consistently.

With modern frozen features, a well-regularised convex probe reaches the best absolute performance, while controlled
comparisons leave little room for more elaborate methods on identical features. DINOv3 makes the limiting case concrete:
without an aligned text head ($\kappa=0$), it is weak at one shot but reaches the reported frontier by eight. Combining
vision-language and self-supervised features is most useful when the aligned encoder lacks strong visual features. Text
therefore matters most when labels are extremely scarce; as visual evidence accumulates, representation quality matters
more than adaptation complexity.

{\small\bibliographystyle{ieeenat_fullname}\bibliography{refs}}

\clearpage
\appendix
\providecommand{\propref}{Proposition~\ref{prop:kappa}}
\providecommand{\taspref}{Eq.~\eqref{eq:tasp}}

\section{Proof of \propref}
\label{app:proof}
Given $m_c$, the $n$ support features of class $c$ are i.i.d.\ $\mathcal N(m_c,\sigma^2 I)$, so their mean is a
sufficient statistic with $\bar x_c\mid m_c\sim\mathcal N(m_c,\frac{\sigma^2}{n}I)$. The text is an independent
observation $t_c\mid m_c\sim\mathcal N(m_c,\tau^2 I)$. With a flat prior on $m_c$, the posterior is proportional to the
product of the two Gaussian likelihoods, hence Gaussian with precision $n/\sigma^2+1/\tau^2$ and mean
\begin{equation*}
\hat m_c=\frac{\frac{n}{\sigma^2}\bar x_c+\frac{1}{\tau^2}t_c}{\frac{n}{\sigma^2}+\frac{1}{\tau^2}}
=\frac{n\,\bar x_c+\kappa\,t_c}{n+\kappa},\qquad \kappa=\frac{\sigma^2}{\tau^2}.
\end{equation*}
The same expression is obtained by assigning $t_c$ a total sample weight $\kappa$ (text as samples) or by using
$\mathcal N(t_c,\tau^2 I)$ as the prior on $m_c$ (text as prior), which is the sense of consequence (iii). With image
covariance $\Sigma$ and text covariance $\Sigma/\kappa$, the plug-in Bayes classifier is the LDA rule
built on $\hat m_c$, \ie a linear classifier; \method{} is its discriminative counterpart. $\square$

\paragraph{Estimating $\kappa$ from the support.}
For $n\geq2$ the method of moments gives $\hat\sigma^2$ as the pooled per-coordinate within-class variance of the shot
embeddings (augmented views averaged) and
$\hat\tau^2=\frac1{CD}\sum_c\|\bar x_c-s\,t_c\|^2-\hat\sigma^2/n$, where $D$ is the feature dimension and $s$ rescales the text
to the average norm of the class means; $\hat\kappa=\hat\sigma^2/\hat\tau^2$. The model assumes isotropic,
class-independent noise for both text and images, with potentially different variances. In contrastive VLMs the image--text residual is concentrated in few directions (the modality
gap~\cite{liang2022mind}), which inflates $\hat\tau^2$ and biases $\hat\kappa$ downwards. Without any validation, $\hat\kappa$ still improves on images only (CLIP: \NumKappaAccImgClip\% with images only, \NumKappaAccHatClip\% with $\hat\kappa$, \NumKappaAccValClip\% at the validation-optimal $\kappa$; 2 to 16 shots).

\section{Experimental details}
\label{app:details}

\paragraph{Shots and splits.}
For the 10 datasets other than ImageNet we use the CoOp split files and the seeded sampler of CoOp/Dassl
(\texttt{random.seed(seed)}, \texttt{random.sample} per class in label order); comparing our DINOv3 features with
features of the support sets used in the literature confirmed identical images. For ImageNet, Dassl lists the training
images with \texttt{os.listdir}, whose order is file-system dependent, so we apply the same sampler to the sorted list. The
benchmark's test split for ImageNet is the 50{,}000-image validation set, which we never use for selection.

\paragraph{Re-implemented methods.}
All methods run on the same pre-processed features (each block centred on the support mean and normalised) and are selected
with the global protocol; grids were extended whenever the selected value fell on the boundary.
\begin{itemize}\itemsep0pt
\item \textbf{Text as prior only (CLAP-like)}~\cite{silva2024clap}: \taspref{} with $w=0$, $\lambda_0\in\{10^{-7},\dots,10\}$.
\item \textbf{Cross-modal LP}~\cite{lin2023crossmodal}: \taspref{} with $w=n$, $W_0=0$, $\lambda_0\in\{10^{-5},10^{-4},10^{-3}\}$.
\item \textbf{Logit ensemble} (WiSE-FT style~\cite{wortsman2022wiseft}): visual probe $+\,\alpha\,s\,\mathrm{ZS}$.
\item \textbf{Tip-Adapter}~\cite{zhang2022tip}: $s\,\mathrm{ZS}+\alpha\exp(-\beta(1-xK^\top))Y$.
\item \textbf{APE}~\cite{zhu2023ape}: channel refinement and three-way relation, training-free.
\item \textbf{ProKeR}~\cite{bendou2025proker}: closed-form kernel ridge regression proximal to the zero-shot predictor.
\item \textbf{GDA}~\cite{wang2024gda}: shared covariance with a shrinkage precision estimator, $s\,\mathrm{ZS}+\alpha\,\mathrm{LDA}$.
\item \textbf{LP++}~\cite{huang2024lp++}: $x(W+\mathrm{diag}(\alpha)T)^\top$, $W$ initialised at the class means.
\item \textbf{AMU-Tuning}~\cite{tang2024amu}: $s_0+\frac{\beta}{\kappa}(Wf_{\text{aux}}+b)$ with $f_{\text{aux}}$ the SSL block, $W$
initialised at its class means, loss $(1-\lambda)\,\ell(Wf_{\text{aux}})+\lambda\,\ell(s)$, AdamW (lr $10^{-3}$, 50 epochs),
$\kappa=\mathrm{kurtosis}(\mathrm{softmax}(s_0))^{\rho}$; $\lambda\in\{0.2,0.35,0.5\}$, $\beta\in\{0.25,\dots,4\}$,
$\rho\in\{0,0.4\}$.
\item \textbf{CaFo}~\cite{zhang2023cafo}: one cache per block, $\hat p_{\text{ZS}}+\alpha\sum_i\mathrm{softmax}_i(\langle\hat
p_i,\hat p_{\text{ZS}}\rangle)\hat p_i$ with z-scored logits, training-free or with keys fine-tuned as in Tip-Adapter-F;
without GPT-3 prompts or DALL-E images.
\item \textbf{CLIP-LoRA}~\cite{zanella2024cliplora}: rank 2, $\alpha=1$, on $q,k,v$ of every layer of both encoders, AdamW
(lr $2\cdot10^{-4}$, weight decay $10^{-2}$), cosine schedule, batch 32, $500\times n$ iterations, random resized crop
(scale 0.08--1) and flip, fp16; LoRA is applied as a parametrisation of the fused $qkv$ projection (no LoRA dropout).
\end{itemize}

\paragraph{Pretraining data of the encoders.}
LVD-142M (DINOv2) retrieves web images close to curated seed sets that include the training splits of ImageNet-1k,
Caltech101, DTD, FGVC-Aircraft, Flowers102, Food101, Oxford Pets, Stanford Cars and SUN397, and removes near-duplicates of
the test and validation sets of benchmarks~\cite{oquab2024dinov2}. LVD-1689M (DINOv3) is also retrieval-based, with seed
datasets that are not listed and no documented de-duplication~\cite{simeoni2025dinov3}. Web-SSL is trained on MetaCLIP's
MC-2B images~\cite{fan2025webssl}, whose pool was de-duplicated against benchmark and ImageNet data~\cite{xu2024metaclip}.

\section{Base-to-novel generalisation}
\label{app:b2n}
\begin{table}[t]\centering\small\setlength{\tabcolsep}{3.5pt}
\resizebox{\columnwidth}{!}{\begin{tabular}{lccc}\toprule
Method & Base & Novel & HM\\\midrule
\multicolumn{4}{l}{\emph{Prior work, CLIP ViT-B/16}}\\
Zero-shot CLIP & 69.3 & 74.2 & 71.7\\
CoOp~\cite{zhou2022coop} & 82.7 & 63.2 & 71.7\\
CoCoOp~\cite{zhou2022cocoop} & 80.5 & 71.7 & 75.8\\
MaPLe~\cite{khattak2023maple} & 82.3 & 75.1 & 78.6\\
PromptSRC~\cite{khattak2023promptsrc} & 84.3 & 76.1 & 80.0\\
MMRL~\cite{guo2025mmrl} & 85.7 & 77.2 & 81.2\\
Best, no extra data (DeAR)~\cite{ma2026dear} & 85.9 & 79.7 & 82.7\\
Best overall (PromptKD+ATPrompt)~\cite{li2025atprompt}$^\ast$ & 87.1 & 81.8 & 84.4\\
Best, CLIP ViT-L/14 (2SFS)~\cite{farina2025twostage} & 89.1 & 79.6 & 84.1\\
\midrule\multicolumn{4}{l}{\emph{Prompt learning on our backbones (our runs)}}\\
CoOp, our run (PE-Core; 3 seeds) & 89.2 & 81.3 & 85.1\\
CoOp, our run (SigLIP; 3 seeds) & 82.5 & 79.7 & 81.1\\
KgCoOp, our run (PE-Core; 3 seeds) & 88.6 & 86.8 & 87.7\\
PromptSRC, our run (PE-Core; 1 valid seed) & 90.2 & 86.3 & 88.2\\
\midrule\multicolumn{4}{l}{\emph{Ours: \method{} on base classes, zero-shot on novel classes}}\\
\method{} CLIP $\|$ DINOv3 & 88.8 & 74.3 & 80.9\\
\method{} SigLIP $\|$ DINOv3 & 90.0 & 80.7 & 85.1\\
\method{} PE-Core & 89.8 & 85.8 & 87.8\\
\method{} PE-Core $\|$ DINOv3 & 90.9 & 85.8 & 88.3\\
\method{} PE-Core-L & 92.7 & \textbf{88.9} & \textbf{90.8}\\
\method{} PE-Core-L $\|$ DINOv3 & \textbf{92.9} & \textbf{88.9} & \textbf{90.8} \\
\bottomrule\end{tabular}}\caption{\textbf{Base-to-novel generalisation} (16 shots on the base half of the classes, 11 datasets; HM of the averaged base and novel accuracies). Novel classes have no image: our novel accuracy is the zero-shot accuracy of the vision-language model. $^\ast$\ uses unlabeled images and/or LLM-generated text. Bold: best in each column.}\label{tab:b2n}\end{table}

Following CoCoOp~\cite{zhou2022cocoop}, each dataset's classes are split in half; the 16-shot support covers base classes
only, and accuracy is measured on base and on novel classes (\cref{tab:b2n}). \method{} uses no image of a novel class,
so on novel classes it falls back to the zero-shot classifier of its aligned block. We also tried predicting novel weights
from base ones through text similarities, and training a single probe in which novel classes only have their class name;
validation selected the zero-shot classifier in both cases. With PE-Core-L, \method{} reaches a harmonic mean
of \NumBtnPeLHm\% (PE-Core$\|$DINOv3: \NumBtnPeDinoHm\%), above the best reported method without extra data
(\NumBtnBestClean\%) and the best with extra data (\NumBtnBestAll\%). Most of this margin is PE-Core-L's zero-shot accuracy
on novel classes (\NumBtnPeLNovel\%), which \method{} leaves untouched; the base classes add the rest
(\NumBtnPeLBase\%). With CLIP's text, novel accuracy is
CLIP's zero-shot accuracy (\NumBtnClipDinoNovel\%), below prompt learners that improve it, and the harmonic mean
(\NumBtnClipDinoHm\%) is below the state of the art: base-to-novel measures the zero-shot quality of the text head.\ifdefined\NumBtnCoopPeNovel{} Prompt learning on
PE-Core (CoOp, our runs) lowers novel accuracy to \NumBtnCoopPeNovel\% (harmonic mean \NumBtnCoopPeHm\%).\fi\ifdefined\NumBtnKgcoopPeNovel{} Prompt learners that regularise towards the hand-crafted prompt do raise it: on
PE-Core (appendix~\ref{app:prompt}), KgCoOp~\cite{yao2023kgcoop} reaches \NumBtnKgcoopPeNovel\% on novel classes and PromptSRC \NumBtnPromptsrcPeNovel\%, above the
zero-shot accuracy that \method{} keeps (\NumBtnPeNovel\%). At this matched backbone PromptSRC is above \method{} in
harmonic mean (\NumBtnPromptsrcPeHm\% vs \NumBtnPeHm\%) and KgCoOp slightly below (\NumBtnKgcoopPeHm\%): when the test
classes are unseen, adapting the text head pays, and \method{} does not adapt it.\fi

\section{Fine-tuning the encoders: CLIP-LoRA}
\label{app:lora}
\begin{table}[t]\centering\small\setlength{\tabcolsep}{3pt}
\resizebox{\columnwidth}{!}{\begin{tabular}{lcccc}\toprule
Method & encoders & back-prop. through & cached features & time / dataset\\\midrule
\method{} (ours) & frozen & no encoder & yes & 0.1\,s (0.1--0.9)\\
CoOp~\cite{zhou2022coop} & frozen & text encoder & image side & 32\,min (2--854)\\
CLIP-LoRA~\cite{zanella2024cliplora} & fine-tuned & both encoders & no & 45\,min (30--110)\\
\bottomrule\end{tabular}}\caption{\textbf{Cost of the three families of methods} (PE-Core, 16 shots, one RTX 4070; median and range over the 10 non-ImageNet datasets; \method{} on PE-Core$\|$DINOv3, after caching the features).}\label{tab:cost}\end{table}

CLIP-LoRA~\cite{zanella2024cliplora} adds rank-2 updates to the attention projections of both encoders and trains them
for $500\times n$ iterations with augmentation. It is a heavier category than \method{}: every iteration back-propagates
through the image and text encoders (\NumLoraParamsMin--\NumLoraParamsMax{}M trained parameters), the adapted embeddings
cannot be cached or shared across tasks, and a 16-shot run takes \NumLoraMinClipSixteen{} minutes per dataset for CLIP
ViT-B/16 and \NumLoraMinPeSixteen{} for PE-Core on an RTX 4070, and \NumLoraMinPeLSixteen{} minutes for PE-Core-L on an
A100 with \NumLoraPeakPeL{} GB of GPU memory; \method{} trains in seconds on cached features. These timing summaries use
the same 10 non-ImageNet datasets on each stated GPU. \Cref{tab:matched} compares the two at matched backbones. Our runs
on CLIP differ by at most 0.7 points from the
originally reported numbers. Fine-tuning keeps a 0.7--2.0 point edge on CLIP alone, whose visual features gain most from
adaptation. On SigLIP and PE-Core-L, the two methods are within 1.0 and 1.2 points, respectively; on PE-Core, \method{}
is 0.6 points below CLIP-LoRA at 1 shot and 0.6--1.6 points above it thereafter. Adding frozen DINOv3 makes \method{}
1.7--4.6 points better than CLIP-LoRA on CLIP and 0.4--2.5 points better on SigLIP. The self-supervised partner also
helps after fine-tuning: concatenating frozen DINOv3 with the CLIP-LoRA-adapted PE-Core features gains 0.1--1.6 points
over CLIP-LoRA alone (seeds 2--3), reaching 83.4\% at 1 shot and 89.0\% at 16 shots, but at a far higher cost than a
probe on frozen features.
\begin{table}[t]\centering\small\setlength{\tabcolsep}{3pt}
\resizebox{\columnwidth}{!}{\begin{tabular}{lccccc}\toprule
Method & 1 & 2 & 4 & 8 & 16\\\midrule
\multicolumn{6}{l}{\emph{CLIP ViT-B/16}}\\
CLIP-LoRA, reported & 72.5 & 75.5 & 77.4 & 80.3 & 83.0\\
CLIP-LoRA, our run & 73.1 & 74.8 & 77.2 & 80.0 & 83.1\\
\method{} CLIP & 71.1 & 73.6 & 76.5 & 78.8 & 81.4\\
\method{} CLIP $\|$ DINOv3 & \textbf{74.8} & \textbf{77.9} & \textbf{81.7} & \textbf{84.6} & \textbf{87.2} \\
\midrule
\multicolumn{6}{l}{\emph{SigLIP ViT-B/16}}\\
CLIP-LoRA, our run & 79.1 & 80.2 & 81.8 & 83.9 & 85.9\\
\method{} SigLIP & 78.1 & 80.0 & 82.0 & 83.7 & 85.4\\
\method{} SigLIP $\|$ DINOv3 & \textbf{79.5} & \textbf{81.7} & \textbf{84.3} & \textbf{86.2} & \textbf{88.3} \\
\midrule
\multicolumn{6}{l}{\emph{PE-Core ViT-B/16}}\\
CLIP-LoRA, our run & 83.3 & 82.9 & 83.8 & 85.6 & 87.4\\
CLIP-LoRA $+$ frozen DINOv3$^\circ$ & \textbf{83.4} & 83.7 & 85.0 & 87.2 & 89.0\\
\method{} PE-Core & 82.7 & 84.1 & 85.4 & 86.6 & 88.0\\
\method{} PE-Core $\|$ DINOv3 & 83.2 & \textbf{84.6} & \textbf{86.1} & \textbf{87.7} & \textbf{89.2} \\
\midrule
\multicolumn{6}{l}{\emph{PE-Core ViT-L/14 (336 px)}}\\
CLIP-LoRA, our run & \textbf{88.3} & 88.0 & 88.8 & 90.2 & 91.4\\
\method{} PE-Core-L & 87.1 & \textbf{88.2} & 89.4 & 90.3 & 91.4\\
\method{} PE-Core-L $\|$ DINOv3 & 87.1 & \textbf{88.2} & \textbf{89.5} & \textbf{90.5} & \textbf{91.6} \\
\bottomrule\end{tabular}}\caption{\textbf{Comparisons at matched backbone, including fine-tuning of the encoders} (11 datasets). CLIP-LoRA changes the embeddings: it back-propagates through both encoders, cannot reuse cached features, and costs up to two hours per dataset instead of seconds (\cref{app:lora}). Our CLIP-LoRA runs use the paper's fixed hyper-parameters (rank 2, all layers, both encoders) on the same splits and shots. $^\circ$\ seeds 2--3, with the configuration selected on the original 10 datasets. Bold: best within each block.}\label{tab:matched}\end{table}

\section{Prompt learners on our backbones}
\label{app:prompt}
CoOp~\cite{zhou2022coop} (PE-Core and SigLIP), KgCoOp~\cite{yao2023kgcoop} and PromptSRC~\cite{khattak2023promptsrc} (PE-Core) are re-run
with the configurations of their official repositories, on the same splits, few-shot samples and base/novel class
halves as \method{}, with three seeds attempted for each run. \Cref{tab:b2n} reports their 11-dataset base-to-novel
aggregates; collapsed PromptSRC runs are handled as described below. We omit these methods from the standard few-shot
comparison because their ImageNet runs were incomplete.
KgCoOp: four context tokens initialised to ``a photo of a'', weight 8 on the
distance to the dataset's hand-crafted prompt, SGD (learning rate 0.002, batch 128, 100 epochs). PromptSRC: deep
vision and language prompts (depth 9, four tokens each), its three self-regularisation losses and Gaussian prompt
aggregation with its per-dataset few-shot schedules (SGD, learning rate 0.0025, batch 4; 20 epochs for base-to-novel,
50 for few-shot). On PE-Core, visual prompts are inserted after the class token and removed before the attention
pooling, so that with zero-depth prompts both models reproduce the frozen encoder exactly (checked to numerical
precision). With its originally reported learning rate and batch of 4, PromptSRC's training is unstable on PE-Core: in
\NumSrcCollapsed{} of its runs the training loss returned to $\ln C$ (chance accuracy), on A100 GPUs as on our local
GPU and in fp16 (the precision of the official code) as in bf16. Following a rule fixed in advance, such a run is re-run
once with the same seed; the \NumSrcExcluded{} runs that collapse again are excluded and the corresponding means are over
the remaining seeds, which favours PromptSRC. Every other run is kept as is, including those whose loss spiked without
collapsing.

\section{What self-supervised features retain: frequency analysis}
\label{app:freq}
Following frequency analyses of image classifiers~\cite{wang2020hfc}, we low-pass or high-pass filter images in the
Fourier domain (cut-off radius $r$ cycles per image), train a 16-shot linear probe on filtered support images and test it
on filtered test images. We report means over all 11 benchmark datasets. With only the high frequencies ($|f|\geq32$), DINOv3
keeps \NumFreqDinoHigh\% accuracy (\NumFreqDinoHighRel\% of its clean accuracy), DINOv2 \NumFreqDinoTwoHigh\%
(\NumFreqDinoTwoHighRel\%), against \NumFreqClipHigh\% (\NumFreqClipHighRel\%) for CLIP and \NumFreqSigHigh\%
(\NumFreqSigHighRel\%) for SigLIP. The vision-language encoders carry less of the fine, high-frequency information that
fine-grained classes depend on, consistent with their documented blind spots~\cite{tong2024eyes} and with the largest
gains of DINOv3 over SigLIP alone, on FGVC-Aircraft (+12.3 points averaged over shot counts) and EuroSAT (+4.7).

\section{Additional results}
\label{app:more}
\Cref{tab:perds} gives per-dataset accuracies of \method{} next to the best reported value for each dataset and shot
count, \cref{tab:samefeatfull} the corresponding 11-dataset comparison at all shot counts and on additional feature unions, \cref{tab:variants} every feature set at every shot count, and \cref{tab:clf} additional classifiers and ablations.
\begin{table*}[t]\centering\small\setlength{\tabcolsep}{3.2pt}
\begin{tabular}{llccccccccccc|c}\toprule
Shots & Method & INet & Cal & Pets & Cars & Flo & Food & Air & SUN & DTD & Euro & UCF & Avg\\\midrule
\multirow{4}{*}{1} & best reported & 75.8 & 95.8 & 94.1 & 73.4 & 93.2 & 89.9 & 33.9 & 73.7 & 62.6 & 83.4 & 81.4 & --\\
& PE-Core-L & \textbf{82.7} & \textbf{98.1} & \textbf{95.9} & \textbf{93.9} & \textbf{96.5} & \textbf{94.0} & \textbf{70.3} & \textbf{77.9} & \textbf{73.6} & \textbf{86.4} & \textbf{89.3} & 87.1\\
& PE-Core $\|$ DINOv3 & \textbf{77.8} & \textbf{97.2} & 94.0 & \textbf{90.5} & \textbf{96.8} & 88.9 & \textbf{56.9} & \textbf{75.4} & \textbf{72.1} & 82.5 & \textbf{82.8} & 83.2\\
& CLIP $\|$ DINOv3 & 68.4 & 95.0 & 90.8 & 72.3 & \textbf{95.7} & 84.6 & \textbf{36.9} & 69.0 & 57.4 & 75.8 & 76.5 & 74.8\\
\midrule
\multirow{4}{*}{2} & best reported & 72.9 & 95.7 & 93.8 & 77.2 & 94.2 & 88.2 & 37.8 & 73.7 & 63.7 & 91.7 & 82.0 & --\\
& PE-Core-L & \textbf{83.1} & \textbf{98.3} & \textbf{95.8} & \textbf{94.1} & \textbf{98.8} & \textbf{94.3} & \textbf{73.1} & \textbf{79.6} & \textbf{74.5} & 87.9 & \textbf{90.6} & 88.2\\
& PE-Core $\|$ DINOv3 & \textbf{78.6} & \textbf{97.4} & \textbf{94.5} & \textbf{90.7} & \textbf{99.0} & \textbf{89.1} & \textbf{60.2} & \textbf{77.5} & \textbf{73.1} & 85.7 & \textbf{84.7} & 84.6\\
& CLIP $\|$ DINOv3 & 70.3 & \textbf{95.7} & 91.1 & \textbf{78.6} & \textbf{98.4} & 85.1 & \textbf{45.1} & 71.6 & 62.7 & 78.4 & 79.8 & 77.9\\
\midrule
\multirow{4}{*}{4} & best reported & 77.2 & 96.8 & 94.3 & 81.2 & 96.8 & 88.0 & 43.8 & 75.9 & 69.4 & 93.1 & 84.8 & --\\
& PE-Core-L & \textbf{83.6} & \textbf{98.6} & \textbf{95.8} & \textbf{94.0} & \textbf{99.5} & \textbf{94.3} & \textbf{75.1} & \textbf{81.3} & \textbf{77.8} & 91.6 & \textbf{91.4} & 89.4\\
& PE-Core $\|$ DINOv3 & \textbf{79.4} & \textbf{98.0} & 94.2 & \textbf{91.3} & \textbf{99.6} & \textbf{89.5} & \textbf{64.2} & \textbf{78.8} & \textbf{76.6} & 89.7 & \textbf{86.1} & 86.1\\
& CLIP $\|$ DINOv3 & 73.0 & \textbf{97.0} & 92.3 & \textbf{84.6} & \textbf{99.5} & 85.9 & \textbf{53.7} & 74.3 & \textbf{69.5} & 86.3 & 82.9 & 81.7\\
\midrule
\multirow{4}{*}{8} & best reported & 74.4 & 96.8 & 94.2 & 87.2 & 98.0 & 88.9 & 55.6 & 76.7 & 73.3 & 91.8 & 88.6 & --\\
& PE-Core-L & \textbf{84.1} & \textbf{98.6} & \textbf{95.7} & \textbf{94.4} & \textbf{99.6} & \textbf{94.4} & \textbf{79.3} & \textbf{82.6} & \textbf{80.0} & \textbf{92.4} & \textbf{92.6} & 90.3\\
& PE-Core $\|$ DINOv3 & \textbf{80.5} & \textbf{98.3} & \textbf{94.7} & \textbf{92.8} & \textbf{99.7} & \textbf{89.8} & \textbf{70.1} & \textbf{80.3} & \textbf{78.8} & 91.5 & 88.0 & 87.7\\
& CLIP $\|$ DINOv3 & \textbf{75.8} & \textbf{97.3} & 93.4 & \textbf{88.8} & \textbf{99.7} & 87.1 & \textbf{63.5} & 76.4 & \textbf{74.5} & 87.9 & 85.8 & 84.6\\
\midrule
\multirow{4}{*}{16} & best reported & 77.8 & 97.4 & 95.3 & 91.1 & 99.4 & 89.3 & 69.0 & 79.8 & 77.2 & 94.4 & 89.9 & --\\
& PE-Core-L & \textbf{84.7} & \textbf{98.7} & \textbf{95.6} & \textbf{95.0} & \textbf{99.8} & \textbf{94.6} & \textbf{82.3} & \textbf{83.8} & \textbf{82.3} & \textbf{94.8} & \textbf{93.5} & 91.4\\
& PE-Core $\|$ DINOv3 & \textbf{81.4} & \textbf{98.4} & 95.0 & \textbf{94.2} & \textbf{99.8} & \textbf{90.3} & \textbf{76.1} & \textbf{81.3} & \textbf{81.5} & 94.1 & 89.3 & 89.2\\
& CLIP $\|$ DINOv3 & \textbf{78.0} & \textbf{97.9} & 94.3 & \textbf{91.8} & \textbf{99.8} & 88.2 & \textbf{71.9} & 78.1 & \textbf{78.2} & 92.7 & 88.1 & 87.2\\
\bottomrule
\end{tabular}\caption{Per-dataset accuracy of \method{}. ``Best reported'': best reported value of any CLIP ViT-B/16 method for that dataset and shot count (an oracle that no single method reaches). Bold: ours $\geq$ best reported.}\label{tab:perds}\end{table*}

\begin{table*}[t]\centering\small\setlength{\tabcolsep}{2.4pt}
\resizebox{\textwidth}{!}{\begin{tabular}{l|ccccc|ccccc|ccccc|ccccc|ccccc}\toprule
 & \multicolumn{5}{c}{PE-Core-L} & \multicolumn{5}{c}{PE-Core $\|$ DINOv3} & \multicolumn{5}{c}{PE-Core-L $\|$ DINOv3} & \multicolumn{5}{c}{SigLIP $\|$ DINOv3} & \multicolumn{5}{c}{CLIP $\|$ DINOv3}\\
Method (same features) & 1 & 2 & 4 & 8 & 16 & 1 & 2 & 4 & 8 & 16 & 1 & 2 & 4 & 8 & 16 & 1 & 2 & 4 & 8 & 16 & 1 & 2 & 4 & 8 & 16\\\midrule
\method{} (ours) & 87.1 & \textbf{88.2} & \textbf{89.4} & \textbf{90.3} & \textbf{91.4} & \textbf{83.2} & \textbf{84.6} & \textbf{86.1} & \textbf{87.7} & \textbf{89.2} & 87.1 & \textbf{88.2} & \textbf{89.5} & \textbf{90.5} & \textbf{91.6} & \textbf{79.5} & 81.7 & \textbf{84.3} & \textbf{86.2} & \textbf{88.3} & \textbf{74.8} & \textbf{77.9} & \textbf{81.7} & \textbf{84.6} & \textbf{87.2}\\
\quad text as samples only & 85.9 & 87.2 & 88.6 & 89.7 & 90.8 & 82.1 & 83.6 & 85.5 & 87.1 & 88.9 & 86.5 & 87.5 & 89.0 & 90.1 & 91.4 & 78.5 & 81.3 & 84.2 & 86.2 & \textbf{88.3} & 74.0 & 77.4 & 81.5 & 84.4 & 87.1\\
\quad text as prior only (CLAP-like~\cite{silva2024clap}) & \textbf{87.2} & 88.1 & 89.3 & \textbf{90.3} & \textbf{91.4} & 83.0 & 84.4 & \textbf{86.1} & \textbf{87.7} & \textbf{89.2} & \textbf{87.2} & 88.0 & 89.4 & \textbf{90.5} & \textbf{91.6} & 79.4 & \textbf{81.8} & \textbf{84.3} & \textbf{86.2} & \textbf{88.3} & 74.3 & 77.6 & \textbf{81.7} & \textbf{84.6} & \textbf{87.2}\\
\quad closed form, least squares (TASC) & 85.3 & 86.1 & 87.7 & 89.0 & 90.1 & 81.4 & 82.8 & 84.7 & 86.1 & 87.5 & 85.8 & 86.8 & 88.5 & 89.4 & 90.5 & 76.8 & 79.4 & 82.6 & 85.0 & 87.2 & 71.8 & 75.6 & 80.2 & 83.4 & 86.1\\
\midrule
CLAP~\cite{silva2024clap} & 84.7 & 86.5 & 87.6 & 88.6 & 89.4 & 80.7 & 82.8 & 84.5 & 85.5 & 86.8 & 85.1 & 86.7 & 88.0 & 89.1 & 90.0 & 75.6 & 78.5 & 80.7 & 82.5 & 84.1 & 70.4 & 74.0 & 77.0 & 79.1 & 81.3\\
Cross-modal LP~\cite{lin2023crossmodal} & 81.0 & 84.4 & 87.3 & 89.1 & 90.7 & 76.6 & 80.5 & 84.0 & 86.5 & 88.6 & 82.7 & 85.0 & 87.6 & 89.4 & 91.0 & 75.5 & 79.4 & 83.3 & 85.9 & 88.1 & 71.2 & 76.2 & 81.1 & 84.4 & \textbf{87.2}\\
LP++~\cite{huang2024lp++} & 83.0 & 85.6 & 88.0 & 89.7 & 91.0 & 77.2 & 81.2 & 84.6 & 87.0 & 88.8 & 82.7 & 85.3 & 88.1 & 89.8 & 91.3 & 75.8 & 79.6 & 83.5 & 85.8 & 88.1 & 71.2 & 76.0 & 81.1 & 84.4 & 87.0\\
ProKeR~\cite{bendou2025proker} & 85.2 & 86.4 & 87.9 & 89.3 & 90.6 & 80.8 & 82.3 & 84.2 & 86.0 & 88.0 & 85.5 & 86.5 & 88.0 & 89.4 & 90.8 & 76.6 & 79.1 & 82.4 & 85.0 & 87.5 & 71.6 & 75.4 & 80.0 & 83.3 & 86.3\\
GDA~\cite{wang2024gda} & 68.9 & 83.1 & 88.5 & 90.2 & 90.2 & 66.1 & 80.4 & 85.7 & 86.9 & 86.4 & 70.8 & 84.5 & 89.0 & 90.1 & 89.8 & 65.2 & 77.3 & 83.6 & 85.6 & 85.5 & 61.7 & 73.6 & 81.2 & 84.8 & 86.7\\
Tip-Adapter~\cite{zhang2022tip} & 84.5 & 85.6 & 86.9 & 88.0 & 88.6 & 79.9 & 81.6 & 83.3 & 84.6 & 85.7 & 84.7 & 85.8 & 87.1 & 88.0 & 88.7 & 75.6 & 77.9 & 80.8 & 83.1 & 84.8 & 70.7 & 74.0 & 77.7 & 80.7 & 83.0\\
APE~\cite{zhu2023ape} & 84.4 & 85.4 & 86.7 & 87.8 & 88.7 & 79.7 & 81.3 & 83.0 & 84.6 & 86.1 & 84.6 & 85.8 & 87.2 & 88.4 & 89.3 & 75.6 & 78.0 & 80.8 & 83.1 & 84.8 & 70.8 & 73.9 & 77.7 & 80.7 & 83.0\\
Logit ensemble & 84.8 & 86.0 & 87.8 & 89.4 & 90.7 & 80.1 & 82.6 & 85.0 & 86.9 & 88.7 & 84.7 & 86.4 & 88.3 & 89.9 & 91.3 & 76.1 & 79.6 & 83.4 & 85.8 & 88.0 & 71.0 & 76.1 & 81.2 & 84.4 & \textbf{87.2}\\
\midrule
AMU-Tuning~\cite{tang2024amu} & -- & -- & -- & -- & -- & 79.9 & 81.5 & 82.6 & 84.5 & 86.1 & 84.0 & 85.0 & 86.0 & 87.2 & 88.1 & 76.1 & 78.6 & 80.8 & 82.7 & 84.6 & 72.6 & 75.3 & 78.2 & 81.3 & 84.0\\
CaFo caches~\cite{zhang2023cafo} & -- & -- & -- & -- & -- & 76.1 & 79.6 & 81.7 & 83.0 & 83.8 & 82.1 & 84.6 & 86.4 & 87.7 & 88.1 & 71.1 & 74.7 & 78.0 & 79.9 & 80.8 & 62.8 & 67.4 & 71.5 & 73.8 & 75.5\\
\quad with fine-tuned keys & -- & -- & -- & -- & -- & 79.2 & 81.9 & 84.1 & 85.8 & 87.3 & 84.3 & 86.2 & 88.2 & 89.4 & 90.5 & 73.5 & 77.3 & 80.9 & 83.2 & 85.4 & 68.6 & 73.3 & 78.1 & 82.1 & 85.4\\
\bottomrule\end{tabular}}\caption{Prior methods on exactly our features, all shot counts (11 datasets; configurations selected on the 10 datasets other than ImageNet). Bold: best (ties included).}\label{tab:samefeatfull}\end{table*}

\begin{table}[h]\centering\scriptsize\setlength{\tabcolsep}{3pt}
\resizebox{\columnwidth}{!}{\begin{tabular}{lccccc}\toprule
Frozen features (\method{}, text as samples) & 1 & 2 & 4 & 8 & 16\\\midrule
\multicolumn{6}{l}{\emph{One vision-language model}}\\
CLIP ViT-B/16 & 68.4 & 71.8 & 75.1 & 77.7 & 80.6\\
MetaCLIP ViT-B/16 & 73.5 & 76.2 & 79.2 & 81.4 & 83.5\\
SigLIP ViT-B/16 & 76.2 & 78.9 & 81.3 & 83.1 & 85.0\\
SigLIP2 ViT-B/16 & 77.9 & 80.3 & 82.9 & 84.5 & 86.2\\
PE-Core ViT-B/16 & 80.7 & 82.5 & 83.9 & 85.5 & 87.0\\
CLIP ViT-L/14 & 75.6 & 78.5 & 81.1 & 83.1 & 85.5\\
SigLIP2 SO400M/14 & 83.5 & 85.8 & 87.6 & 88.7 & 90.1\\
PE-Core ViT-L/14 (336 px) & 85.9 & 87.2 & 88.6 & 89.7 & 90.8\\
\multicolumn{6}{l}{\emph{Two vision-language models}}\\
CLIP $\|$ SigLIP & 77.4 & 79.5 & 81.6 & 83.4 & 85.5\\
SigLIP $\|$ MetaCLIP & 78.5 & 80.4 & 82.7 & 84.3 & 86.1\\
SigLIP2 $\|$ PE-Core & 82.1 & 83.6 & 85.1 & 86.5 & 88.1\\
\multicolumn{6}{l}{\emph{Vision-language + self-supervised}}\\
CLIP $\|$ DINOv2 & 72.4 & 75.9 & 79.5 & 82.4 & 85.5\\
CLIP $\|$ DINOv3 & 74.0 & 77.4 & 81.5 & 84.4 & 87.1\\
MetaCLIP $\|$ Web-SSL$^\ddagger$ & 75.1 & 77.9 & 81.0 & 83.5 & 85.9\\
MetaCLIP $\|$ DINOv2 & 76.0 & 78.7 & 81.8 & 84.2 & 86.6\\
MetaCLIP $\|$ DINOv3 & 77.0 & 79.6 & 83.0 & 85.4 & 87.6\\
SigLIP $\|$ Web-SSL$^\ddagger$ & 77.4 & 79.9 & 82.4 & 84.6 & 86.9\\
SigLIP $\|$ DINOv2 & 77.8 & 80.4 & 83.0 & 85.1 & 87.3\\
SigLIP $\|$ DINOv3 & 78.5 & 81.3 & 84.2 & 86.2 & 88.3\\
SigLIP2 $\|$ DINOv3 & 79.8 & 82.4 & 85.1 & 86.8 & 88.6\\
PE-Core $\|$ DINOv3 & 82.1 & 83.6 & 85.5 & 87.1 & 88.9\\
CLIP-L/14 $\|$ DINOv3 & 78.6 & 81.5 & 84.6 & 86.7 & 88.7\\
PE-Core-L $\|$ DINOv3 & \textbf{86.5} & \textbf{87.5} & \textbf{89.0} & \textbf{90.1} & \textbf{91.4} \\
\multicolumn{6}{l}{\emph{Vision-language + other partners}}\\
SigLIP $\|$ DINO v1$^\S$ & 77.0 & 79.5 & 82.1 & 84.0 & 86.0\\
PE-Core $\|$ DINO v1$^\S$ & 80.1 & 82.4 & 84.4 & 85.9 & 87.6\\
SigLIP $\|$ MAE$^\S$ & 72.8 & 75.0 & 78.1 & 81.4 & 84.3\\
PE-Core $\|$ MAE$^\S$ & 76.9 & 77.9 & 80.5 & 83.7 & 86.3\\
SigLIP $\|$ supervised IN-21k & 77.3 & 79.6 & 82.3 & 84.2 & 86.1\\
PE-Core $\|$ supervised IN-21k & 80.3 & 82.6 & 84.6 & 86.3 & 87.8\\
\multicolumn{6}{l}{\emph{Partner only (no text)}}\\
DINOv2 ViT-B/14 & 57.6 & 66.2 & 73.7 & 79.0 & 83.6\\
Web-SSL ViT-L/14 & 51.3 & 61.8 & 70.5 & 77.0 & 82.0\\
DINOv3 ViT-B/16 & 62.0 & 71.3 & 78.9 & 83.1 & 86.2\\
DINO v1 ViT-B/16$^\S$ & 42.9 & 52.2 & 61.2 & 68.3 & 74.2\\
MAE ViT-B/16$^\S$ & 12.1 & 19.6 & 29.4 & 41.3 & 52.0\\
supervised IN-21k ViT-B/16 & 51.7 & 60.0 & 67.8 & 73.0 & 77.6\\
\bottomrule\end{tabular}}\caption{All feature sets, all shot counts (11 datasets; configurations selected globally on the 10 datasets other than ImageNet; text as samples). $^\ddagger$\ Web-SSL DINO is trained on MetaCLIP's images without retrieval curation. $^\S$\ self-supervised on ImageNet-1k only. Bold: best in each column.}\label{tab:variants}\end{table}

\begin{table}[h]\centering\small\setlength{\tabcolsep}{3pt}
\resizebox{\columnwidth}{!}{\begin{tabular}{lccccc}\toprule
PE-Core-L, text + images & 1 & 2 & 4 & 8 & 16\\\midrule
\method{} (text as samples and prior) & \textbf{87.1} & \textbf{88.2} & \textbf{89.4} & \textbf{90.3} & \textbf{91.4}\\
\method{}$^\ast$ & 85.9 & 87.2 & 88.6 & 89.7 & 90.8\\
\midrule
Nearest class mean (posterior means, Prop.~1) & 85.5 & 86.7 & 87.7 & 88.3 & 88.7\\
LDA (shrinkage) & 85.2 & 87.2 & 88.9 & 90.2 & 91.3\\
$k$-NN vote & 75.6 & 76.4 & 78.9 & 81.3 & 83.9\\
Soft nearest neighbours & 85.4 & 86.0 & 87.2 & 87.9 & 88.6\\
MLP (1 hidden layer) & 85.2 & 86.8 & 88.6 & 89.9 & 91.2\\
\method{}$^\ast$, 1 view per shot (vs 10) & 85.7 & 87.0 & 88.4 & 89.5 & 90.7\\
\method{}$^\ast$, text centred on images & 86.6 & 87.8 & 88.9 & 90.0 & 91.0\\
\bottomrule\end{tabular}}\caption{Other classifiers on the union of text and images, and two ablations (11 datasets; configurations selected globally on the 10 datasets other than ImageNet). $^\ast$\ text as samples only. Bold: best in each column.}\label{tab:clf}\end{table}

\paragraph{Transductive extension.}
All main-paper results are inductive: each test image is classified independently, without access to the unlabeled test
pool. For completeness, \cref{tab:trans} reports a transductive extension, denoted \method{}-T, that applies soft $k$-means
to the test pool, initialised by \method{}. Its iterations alternate
\begin{align}
m_c &\propto \frac{1}{V}\sum_{i\in\mathcal S_c}\sum_{r=1}^{V}x_{cir}
       + \sum_j q_{jc}z_j, \\
q_{jc} &= \operatorname{softmax}_c\!\left(\beta\langle z_j,m_c\rangle
          + \log p^{\mathrm{TASP}}_{jc}\right),
\end{align}
where $\mathcal S_c$ contains the support images of class $c$, $x_{cir}$ is the feature of augmented view $r$ of support
image $i$, $V=10$ is the number of views per image, $z_j$ denotes an unlabeled test feature, and $m_c$ is
$\ell_2$-normalised after each update. The assignments are
initialised with the tempered inductive posterior,
$q^{(0)}=\operatorname{softmax}(\ell/t_0)$, where $\ell$ denotes the TASP logits. Hyper-parameters are selected on the
validation pools (\NumTrSelPeL{} for PE-Core-L and \NumTrSelPeDino{} for PE-Core$\|$DINOv3), then fixed on the test
split. The transductive update gains \NumTrGainPeLOne{} and \NumTrGainPeDinoOne{} points, respectively,
at 1 shot, but no longer changes accuracy at 16 shots. These results are reported separately because transductive
methods use information unavailable in the standard inductive protocol~\cite{zanella2024transclip}.
\begin{table}[h]\centering\small\setlength{\tabcolsep}{3pt}
\resizebox{\columnwidth}{!}{\begin{tabular}{lccccc}\toprule
Method & 1 & 2 & 4 & 8 & 16\\\midrule
\multicolumn{6}{l}{\emph{Reported transductive methods, CLIP ViT-B/16, 11 datasets}}\\
ProGrad + TransCLIP-ZS~\cite{zanella2024transclip} & 74.3 & -- & 77.7 & -- & 81.1\\
TIP-Adapter-F + TransCLIP-ZS~\cite{zanella2024transclip} & 75.1 & -- & 77.9 & -- & 81.1\\
GTA-CLIP~\cite{saha2025gtaclip} & 76.1 & -- & 79.5 & -- & 81.4\\
TransCLIP-LoRA~\cite{baklouti2025limo} & -- & 75.3 & 77.2 & -- & 81.9\\
LIMO~\cite{baklouti2025limo} & -- & 77.8 & 80.2 & -- & 84.9\\
Best reported (any protocol), per shot & 76.1 & 77.9 & 80.6 & 83.8 & 86.8\\\midrule
\multicolumn{6}{l}{\emph{Ours, 11 datasets}}\\
\method{} PE-Core-L (inductive) & 87.1 & 88.2 & 89.4 & 90.3 & \textbf{91.4} \\
\method{}-T PE-Core-L (transductive) & \textbf{88.1} & \textbf{88.7} & \textbf{89.6} & \textbf{90.4} & \textbf{91.4} \\
\method{} PE-Core $\|$ DINOv3 (inductive) & 83.2 & 84.6 & 86.1 & 87.7 & 89.2\\
\method{}-T PE-Core $\|$ DINOv3 (transductive) & 84.6 & 85.3 & 86.5 & 87.8 & 89.2\\
\bottomrule\end{tabular}}\caption{\textbf{Transductive extension} (soft $k$-means seeded by \method{}; hyper-parameters selected globally on the validation pools of the 10 non-ImageNet datasets, never on test). Bold: best in each column.}\label{tab:trans}\end{table}

\end{document}